\documentclass[runningheads]{llncs}
\usepackage{graphicx}

\usepackage{amsmath,amssymb} 
\usepackage{color}
\usepackage{hyperref}

\usepackage{tabularx,booktabs}
\newcommand{\beginsupplement}{%
	\setcounter{table}{0}
	\renewcommand{\thetable}{S\arabic{table}}%
	\setcounter{figure}{0}
	\renewcommand{\thefigure}{S\arabic{figure}}%
}

\begin{document}
\title{Toward Scale-Invariance and Position-Sensitive Region Proposal Networks} 

\titlerunning{Toward Scale-Inv. and Pos.-Sens. Reg. Propos. Net.}
%
\author{Hsueh-Fu Lu\and
Xiaofei Du \and
Ping-Lin Chang}
\authorrunning{H.-F. Lu, X. Du and P.-L. Chang}
%

\institute{Umbo Computer Vision \\
\email{\{topper.lu, xiaofei.du, ping-lin.chang\}@umbocv.com}\\
\url{https://umbocv.ai}
}

%
\maketitle              
\begin{abstract}
Accurately localising object proposals is an important precondition for high detection rate for the state-of-the-art object detection frameworks. The accuracy of an object detection method has been shown highly related to the average recall (AR) of the proposals. In this work, we propose an advanced object proposal network in favour of translation-invariance for objectness classification, translation-variance for bounding box regression, large effective receptive fields for capturing global context and scale-invariance for dealing with a range of object sizes from extremely small to large. The design of the network architecture aims to be simple while being effective and with real-time performance. Without bells and whistles the proposed object proposal network significantly improves the AR at 1,000 proposals by $35\%$ and $45\%$ on PASCAL VOC and COCO dataset respectively and has a fast inference time of 44.8 ms for input image size of $640^{2}$. Empirical studies have also shown that the proposed method is class-agnostic to be generalised for general object proposal.


\end{abstract}
\section{Introduction}
Object detection has been a challenging task in computer vision~\cite{Everingham2010,Lin2014}. Significant progress has been achieved in the last decade from traditional sliding-window paradigms~\cite{Felzenszwalb2010,Viola2004} to recent top-performance proposal-based~\cite{Uijlings2013} detection frameworks~\cite{Girshick2015,Girshick2014,He2014}. A proposal algorithm plays a crucial role in an object detection pipeline. On one hand, it speeds up the detection process by considerably reducing the search space for image regions to be subsequently classified. On the other hand, the average recall (AR) of the object proposal method has been shown notably correlating with the precision of final detection, in which AR essentially reveals how accurate the detected bounding boxes are localised comparing with the ground truth~\cite{Hosang2016}.

Instead of using low-level image features to heuristically generate the proposals~\cite{Uijlings2013,Zitnick2014}, trendy methods extract high-level features by using deep convolutional neural networks (ConvNets)~\cite{He2016,Simonyan2014,Zeiler2014} to train a class-agnostic classifier with a large number of annotated objects~\cite{Kuo2015,Pinheiro2016,Ren2015}. For general objectness detection, such supervised learning approaches make an important assumption that given enough number of different object categories, an objectness classifier can be sufficiently generalised to unseen categories. It has been shown that learning-based methods indeed tend to be unbiased to the dataset categories and learn the union of features in the annotated object regions~\cite{Chavali2016,Hosang2016,Kuo2015,Pinheiro2015}. Despite their good performance~\cite{Dai2016,He2016,Ren2015}, there is still much room to improve the recall especially for small objects and accuracy for the bounding box localisation~\cite{Bell2016,Kong2016,Lin2017,Pinheiro2015}.

To tackle object detection using ConvNets at various scales and for more accurate localisation, prior works adopted an encoder-decoder architecture with skip-connections~\cite{Ronneberger2015} for exploiting low-resolution strong semantic and high-resolution weak semantic features~\cite{Lin2017}, used position sensitive score maps for enhancing translation variance and invariance respectively for localisation and classification~\cite{Dai2016}, and used a global convolutional network (GCN) component for enlarging valid receptive field (VRF) particularly for capturing larger image context~\cite{Peng2017}.

In this paper, we devise an advanced object proposal network which is capable of handling a large range of object scales and accurately localising proposed bounding boxes. The proposed network architecture embraces fully convolutional networks (FCNs)~\cite{Long2015} without using fully-connected and pooling layers to preserve spatial information as much as possible. The design takes simplicity into account, in which the features extracted by ConvNets are entirely shared with a light-weight network head as shown in Fig.~\ref{fig:network-chart}. 

Ablation studies have been conducted to show the effectiveness of each designed component. We have empirically found that GCN and position-sensitivity structure can each individually improves the AR at 1,000 proposals. As shown in Table~\ref{tab:pascal-results} and \ref{tab:coco-results}, evaluating the baseline model on PASCAL VOC and COCO dataset, GCN brings performance gains from 0.48 and 0.42 to 0.59 (22\%) and to 0.54 (29\%) respectively, and, the use of position-sensitivity to 0.61 (26\%) and to 0.45 (6\%) respectively. Using them together can furthermore boost the scores to 0.65 (35\%) and to 0.61 (44\%) respectively. Together the proposed framework achieves the state of the art and has a real-time performance.

\section{Related works}
\label{sec:related-works}
Traditional object proposal methods take low-level image features to heuristically propose regions containing objectness. Methods such as Selective Search~\cite{Uijlings2013}, CPMC~\cite{Carreira2012} and MCG~\cite{Arbelaez2014} adopt grouping of multiple hierarchical segmentations to produce the final proposals. Edge boxes~\cite{Zitnick2014} on the other hand takes an assumption that objectness is supposed to have clearer contours. Hosang et al.~\cite{Hosang2016} have comprehensively evaluated different proposal methods. Learning-based proposal approaches have gained more attentions recently. DeepBox~\cite{Kuo2015} uses convolutional neural network to re-rank object proposals based on other bottom-up non-learning proposal methods. Faster R-CNN~\cite{Ren2015} trains a region proposal network (RPN) on a large number of annotated ground truth bounding boxes to obtain high-quality box proposals for object detection. Region-based fully-convolutional network (R-FCN)~\cite{Dai2016} introduces position-sensitive score maps to improve localisation of the bounding boxes at detection stage. Feature pyramid network (FPN)~\cite{Lin2017} takes multi-scale feature maps into account to exploit scale-invariant features for both object proposal and detection stages. Instead of using feature pyramids and learning from ground truth bounding boxes, DeepMask~\cite{Pinheiro2015} and SharpMask~\cite{Pinheiro2016} use feed-forward ConvNets trained by ground truth masks and exploit multi-scale input images to perform mask proposals to achieve state-of-the-art performances.
\begin{figure}[t]
	\centering
	\includegraphics[width=\textwidth]{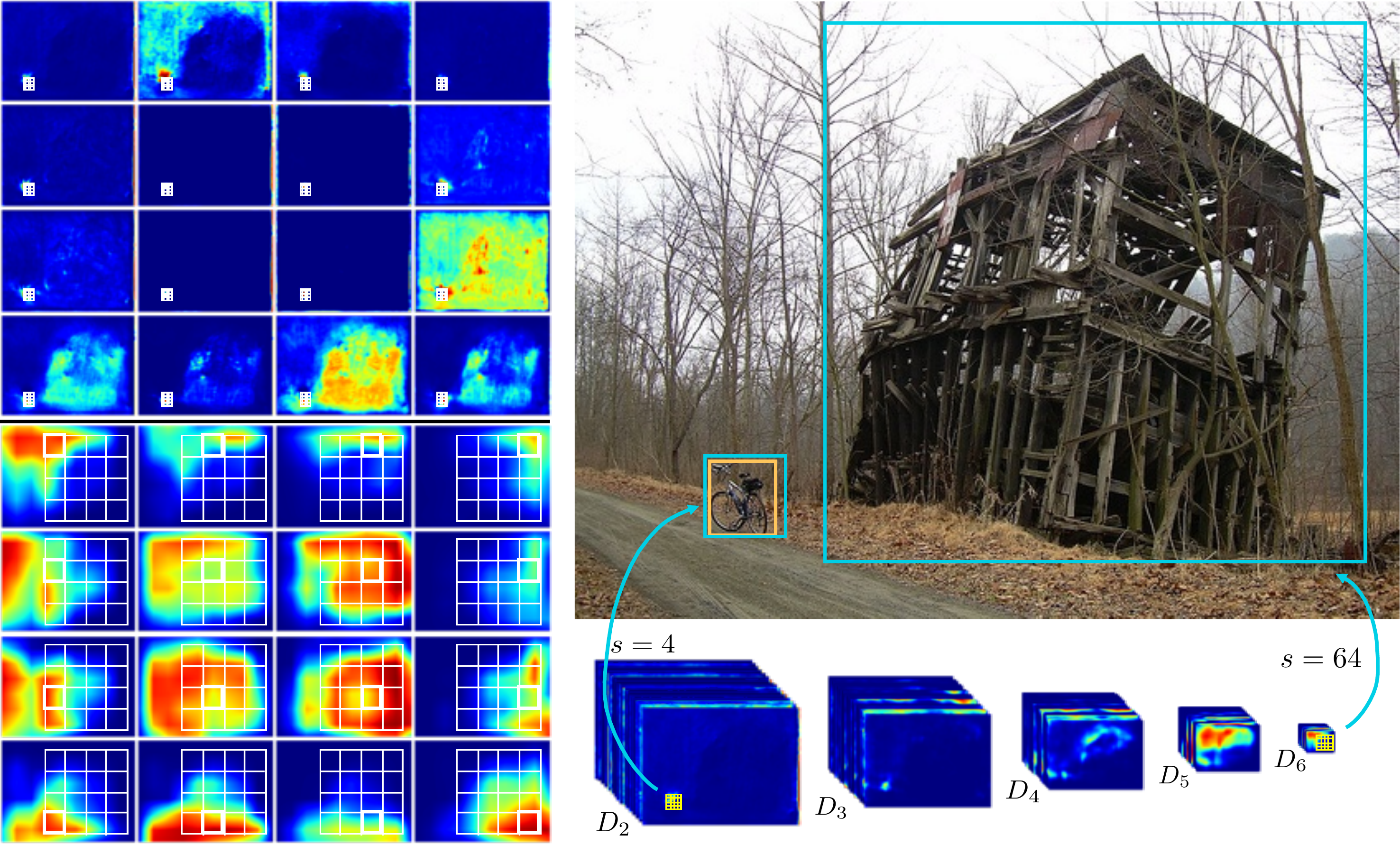}
	\caption{The proposed method. \textbf{Left:} The position-sensitive score maps and windows with $k^2$ grids ($k=4$) in yellow at $D_{2}$ and $D_{6}$ shown at the top and bottom respectively. One can see the $D_{2}$ activates on small objects while $D_{6}$ activates on extreme large ones. Note that $D_{6}$ is enlarged for visualisation, in which the window size is in fact identical to the one shown in $D_{2}$. \textbf{Right:} The windows are mapped into anchors in cyan in the input image with sizes of the multiple of layer stride $s$. \textbf{Both:} The bounding box in orange is the only labeled ground truth (in \texttt{bike} category) on this image from PASCAL VOC 2007. The large object on the right has no ground truth but the proposed class-agnostic method can still be generalised to extract its objectness features as shown in the visualised $D_{6}$ feature maps.}
\label{fig:pyramid-feature-anchor}
\end{figure}

\section{Proposed method}
\label{sec:proposed-method}
Inspired by FPN~\cite{Lin2017} and R-FCN~\cite{Dai2016}, the proposed object proposal method is devised in favour of scale-invariance and position-sensitivity to retain both invariance and variance on translation for respectively classifying and localising objects. We also take VRF into account to learn objectness from a larger image spatial context~\cite{Peng2017}. In addition, instead of regressing and classifying a set of anchors using default profile (i.e., scale and aspect ratio) by a fixed ($3\times3$) convolutional kernel in certain layers~\cite{Lin2017,Ren2015}, we propose directly mapping anchors from sliding windows in each decoding layer together with sharing the position-sensitive score maps.
\begin{figure}[t]
	\centering
	\includegraphics[width=\textwidth]{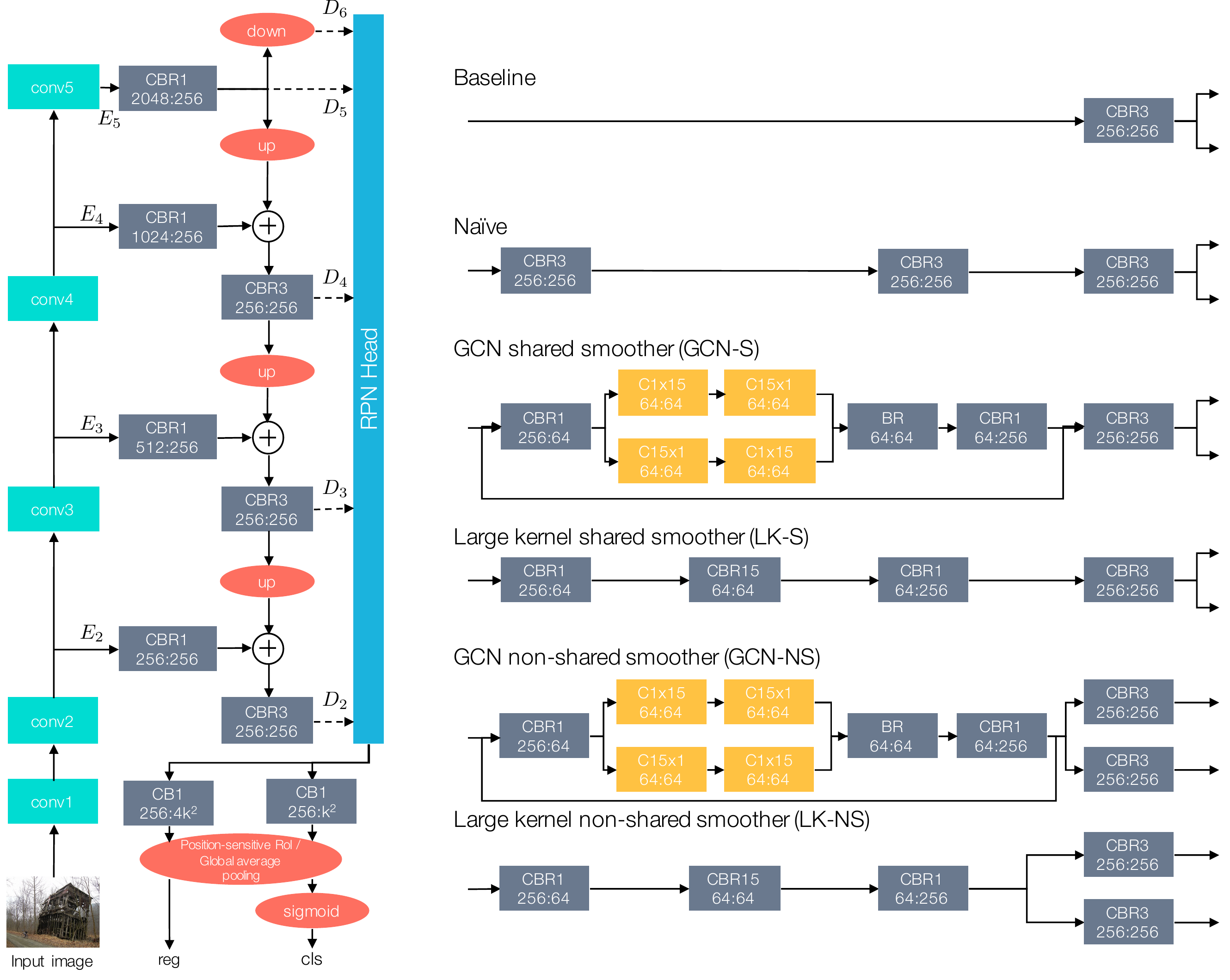}
	\caption{The overall proposed system architecture. \textbf{Left:} The ResNet together with the feature pyramid structure form the general backbone of RPN heads. \textbf{Right:} The structures of different RPN heads. \textbf{Both:} Rectangles are components with learnable parameters to train and ellipses are parameter-free operations. Dash arrow indicates that the RPN head is shared by all feature pyramid levels.}
	\label{fig:network-chart}
\end{figure}
The overall ConvNets takes an input image with arbitrary size to bottom-up encode and top-down decode features with skip connections to preserve object locality~\cite{Ronneberger2015}. Scale-invariance as one of the important traits of the proposed method is thus achieved by extracting multi-scale features from the input image. These semantically weak to strong features are then feed into a series of decoding layers being shared by a RPN head. Anchors are generated by a dense sliding window fashion shared by a bank of position sensitive score maps. In the end, the network regresses the anchors to localise objects ($reg$ for short) and classifies the objectness with scores ($cls$ for short). 

\subsection{Encoder}
\label{sec:encoder}
The encoder is a feed-forward ConvNet as the backbone feature extractor, which scales down by a factor of 2 several times. Although the proposed method can be equipped with any popular ConvNet architectures~\cite{Simonyan2014,Zeiler2014} for the backbone, ResNet~\cite{He2016} is adopted particularly for its FCN structure being able to retain the local information as much as possible.
ResNets are structured with residual \textit{blocks} each consisting of a subset of ConvNets. We note the conv2, conv3, conv4 and conv5 blocks from the original paper~\cite{He2016} as $\{E_{2}, E_{3}, E_{4}, E_{5}\}$ with the corresponding dense sliding window strides $s = \{4, 8, 16, 32\}$ in regard to the input image.

\subsection{Decoder}
\label{sec:decoder}
The decoder recovers the feature resolution for the strongest semantic feature maps from low to high with skip connections in between the corresponding encoder and decoder layers. The skip connection is substantial for the accuracy of bounding box proposal as it propagates detail-preserving and position-accurate features from the encoding process to the decoded features which are later shared by the RPN head.

Specifically with ResNet, the decoding process starts from $E5$ using $1\times1$ convolution and $256$ output channels for feature selection followed by batch normalisation (BN) and rectified linear unit (ReLU) layers, which together we brief as CBR$\{\cdot\}$ where $\cdot$ is the kernel size. Likewise, each skip connection at a layer takes a CBR1 with $256$ output channels. The bottom-up decoding process is therefore done by using bilinear upsampling followed by element-wise addition with the CBR1 selected features from the encoding layers. A CBR3 block is inserted in each decoding layer right after the addition for the purpose of de-aliasing. We note the decoding layers as $\{D_{2}, D_{3}, D_{4}, D_{5}\}$ corresponding to $\{E_{2}, E_{3}, E_{4}, E_{5}\}$ in the encoding layers. An extra $D_{6}$ is added by directly down sampling $D_{5}$ for gaining an even larger stride $s=64$, which is in favor of extremely large objectness detection.

\subsection{RPN heads}
\label{sec:rpn-heads}
A RPN head is in charge of learning features across a range of scales for $reg$ and $cls$. The learnable parameters in the RPN head share all features in the decoding layers to capture different levels of semantics for various object sizes. We will show that the design of a RPN head has a significant impact on the final proposal accuracy in Sec.~\ref{sec:empirical-studies}.
We show a number of different RPN head designs in Fig.~\ref{fig:network-chart}. Each head takes 256 channel feature map as input and outputs two sibling CB1 blocks for \textit{reg} and \textit{cls} with $4 \times k^2$ and $k^2$ channels respectively, where $k^2$ is the number of regular grids for position-sensitive score maps described in Sec.~\ref{position-sensitive-anchors}. 
We regard the state-of-the-art RPN used in FPN~\cite{Lin2017} as a \textit{Baseline} method, in which a CBR3 block is adopted for fusing multi-scale features. Our \textit{Baseline} implementation, which is a bit different from~\cite{Lin2017}, uses BN and ReLU which have been found helpful in converging the end-to-end training.

Inspired by GCN within residual structure~\cite{Peng2017}, we hypothesise that enlarging VRF to learn from larger image context can improve the overall object proposal performance. For the \textit{GCN shared smoother} (GCN-S) and \textit{Large kernel shared smoother} (LK-S), a larger convolution kernel ($15\times15$) is inserted before the CBR3 smoothing. Additionally their non-shared smoother counterpart (GCN-NS and LK-NS) are also compared.

To study the effect of model capacity and the increased number of parameters, a \textit{Na\"ive} head is taken into account, which is simply added with more CBR3 blocks to approximately match the number of learnable parameters compared with other RPN heads. Table~\ref{tab:model-param-time} lists the number of parameters of all RPN heads. Compared with the \textit{Baseline}, the numbers of parameter ratio of the other models are within a $0.015$ standard deviation.

\subsection{Position-sensitive anchors}
\label{position-sensitive-anchors}
We argue that using a default set of scales and aspect ratios to map anchors from a constant-size convolution kernel can potentially undermine the accuracy of \textit{reg} and \textit{cls}. This could be due to the mismatch of the receptive field of network and the mapped anchors. Prior works have used such strategy~\cite{Dai2016,Lin2017,Ren2015} with little exploration of other varieties. To improve the fidelity of relationship between features and anchors with respect to the receptive field of ConvNets, in the proposed method, at each layer, the size of an anchor is calculated by $(w \cdot s) \times (h \cdot s)$ where $w$ and $h$ are the width and height of the sliding window. 

Since the anchor and the sliding window are now naturally mapped, position-sensitive score maps can be further exploited for improving the accuracy of localisation. Fig.~\ref{fig:pyramid-feature-anchor} illustrates the stack of score maps for $k^2$ regular grids in the sliding window. Each grid in the window takes average of its coverage on the corresponding score map (i.e., average pooling). All $k^2$ grids then undergo a global average pooling to output 4-channel $t$ and 1-channel $o$ for the final \textit{reg} and \textit{cls} result respectively. We further feed $o$ to an activation function $sigmoid$ for evaluating the objectness score. Details of position-sensitive pooling can be found in~\cite{Dai2016}. In this paper we use $k=4$ for illustration as well as for all experiments.

\subsection{Implementation details}
\label{sec:implementation-details}
In this paper, all the experiments were conducted with ResNet-50 with the removal of average pooling, fully-connected and softmax layers in the end of the original model. We do not use conv1 in the pyramid due to the high memory footage and too low-level features which contribute very little for semantically representing objectness. The architecture is illustrated in Fig.~\ref{fig:network-chart}.
A set of window sizes $w:h = \{8:8,\, 4:8,\, 8:4,\, 3:9,\, 9:3\}$ are used for the dense sliding windows at each layer for generating anchors. At the most top $D_{6}$ and bottom $D_{2}$ layer, additional window sizes $\{12:12,\, 6:12,\, 12:6,\, 12:4,\, 4:12\}$ and $\{4:4,\, 2:4,\, 4:2\}$ are respectively used for discovering extremely large and small objectness. 

The proposed position-sensitive anchors mapped from the windows are all inside the input image, but the bounding boxes regressed from anchors can possibly exceed the image boundary. We simply discard those bounding boxes exceeding the image boundary. In addition, the number of total anchors depends on the size of input image and the used anchor profile. The effect of anchor number is discussed in the supplementary material.

\subsubsection{Training}
\label{sec:training}
In each image, a large amount of anchors are generated across all decoding layers to be further assigned positive and negative labels. An anchor having intersection-over-union (IoU) with any ground truth bounding box greater than 0.7 is assigned a positive label $p$ and less than 0.3 a negative label $n$. For each ground truth bounding box, the anchor with the highest IoU is also assigned to a positive label, only if the IoU is greater than 0.3. This policy is similar to~\cite{Ren2015} but with the additional lower bound for avoiding distraction of outliers. $N_A$ anchors (half positive and half negative anchors) are selected for each training iteration.
The model can be trained end-to-end with $N_B$ mini-batch images together with the sampled anchors using a defined loss:

\begin{align}
L &= \frac{1}{N_B \cdot N_A}\sum_{i=1}^{N_B}\bigg[\sum_{j=1}^{N_A}\Big[L_{reg}(t_{i,j}^{p}, t_{i,j}^{*}) + L_{cls}(o_{i,j}^{p})\Big] + \sum_{j=1}^{N_A} L_{cls}(o_{i,j}^{n})\bigg],
\end{align}

\noindent where $t$ is the regressed bounding box with $t^*$ as its ground truth correspondent, and $o$ is the objectness score. $L_{reg}$ is the smooth $L_{1}$ loss taking the difference between normalised bounding box coordinates with the ground truth as defined in~\cite{Girshick2015}, and $L_{cls}$ the cross-entropy loss. We use stochastic gradient descent (SGD) with momentum of $0.9$, weight decay of $10^{-4}$ and exponential decay learning rate $l_{e} = l_{0}b^{-\lambda e}$, in which the $e$ is the epoch number and we set $l_{0}=0.1$ and $\lambda=0.1$ for the base $b=10$.

\section{Empirical studies}
\label{sec:empirical-studies}
We have conducted comprehensive experiments for comparing different RPN heads as well as ablation studies to show the impact of position-sensitive score maps. The experiment platform is equipped with an Intel(R) Xeon(R) CPU E5-2650 v4@2.20GHz CPU and Nvidia Titan X (Pascal) GPUs with 12 GB memory. Such hardware spec allowed us to train the models with batch size $N_{B}$ listed in Table~\ref{tab:model-param-time}. 
Note that we particularly focus on conducting ablation studies on different components. In all experiments we therefore did not exploit additional tricks for boosting the performance such as using multi-scale input images for training~\cite{He2014} and testing~\cite{He2016}, iterative regression~\cite{Gidaris2015}, hard example mining~\cite{Shrivastava2016}, etc.
\begin{table}[t]
\caption{The number of parameters in the different models and the corresponding inference time T in $ms$ averaged on the number of all testing images}
\label{tab:model-param-time}
\resizebox{\textwidth}{!} {\begin{tabular}[t!]{llcccclcccc}
\toprule
\multicolumn{2}{c}{} & 
\multicolumn{4}{c}{w/o position-sensitive} & 
\multicolumn{1}{c}{} & 
\multicolumn{4}{c}{w/  position-sensitive} \\
\cmidrule(lr){2-6}
\cmidrule(lr){7-11}
\multicolumn{2}{c}{} &
\multicolumn{1}{c}{\# params} &
\multicolumn{1}{c}{$N_{B}$} &
\multicolumn{1}{c}{$\textrm{T}_{\texttt{07test}}$} &
\multicolumn{1}{c}{$\textrm{T}_{\texttt{minival}}$} &
\multicolumn{1}{c}{} &
\multicolumn{1}{c}{\# params} &
\multicolumn{1}{c}{$N_{B}$} &
\multicolumn{1}{c}{$\textrm{T}_{\texttt{07test}}$} &
\multicolumn{1}{c}{$\textrm{T}_{\texttt{minival}}$} \\
\midrule
Baseline & & 26,858,334 & 28 & 26.6 & 58.2 & & 26,875,104 & 18 & 35.7 & 79.5 	\\ \midrule
Na\"ive  & & 28,039,006 	& 18 & 32.2 & -    & & 28,055,776 & 14 & 41.5 & -	 	\\
GCN-S    & & 27,137,630 & 18 & 34.3 & 76.3 & & 27,154,400 & 14 & 44.1 & 96.1 	\\
LK-S 	 & & 27,81,3470 & 18 & 45.2 & -    & & 27,830,240 & 14 & 55.1 & -	 	\\
GCN-NS 	 & & 27,727,966 & 16 & 35.9 & 83.5 & & 27,744,736 & 12 & 44.8 & 103.6 	\\
LK-NS 	 & & 28,403,806 & 16 & 48.8 & -    & & 28,420,576 & 12 & 57.5 & - 		\\
\bottomrule
\end{tabular}%
}
\end{table}

\subsubsection{Baseline model}
The implementation of our \textit{Baseline} model, with or without using position-sensitive score maps, differ from the original FPN~\cite{Lin2017} in the use of BN and ReLU in the RPN head, as well as the de-aliasing CBR3 block in each layer. In addition, the evaluation in their paper was conducted with rescaling image short side to $800$ pixels. The rest of setting such as anchor generation, the number of pyramid layers, etc. are remained the same. Note that such discrepancy do not affect the ablation studies here to compare the baseline architecture. The main purpose is to assess performance gains when adding other network components.

\subsubsection{Evaluation protocol}
All models are evaluated on PASCAL VOC~\cite{Everingham2010} and COCO~\cite{Lin2014}. For Pascal VOC we used all train and validation dataset in 2007 and 2012 (denoted as \texttt{07+12}), which has in total 16,551 images with 40,058 objects, and report test result on the 2007 test dataset (denoted as \texttt{07test}) consisting of 4,952 images with 12,032 objects. For COCO we employed the union of train and a subset of validation set for in total 109,172 images and 654,212 objects (denoted as \texttt{trainval35k}), and report test results on the rest of validation set for 4,589 images and 27,436 objects (denoted as \texttt{minival}). Our evaluation is consistent with the official COCO evaluation protocol, in which areas marked as "crowds" are ignored and do not affect detector's scores~\cite{Lin2014}. 

In order to perform mini-batch training, we rescaled images in \texttt{07+12} with the long side fixed and zero-pad along the rescaled short side to $640\times640$ for batching, and for images in \texttt{trainval35k}, the short side were fixed to $768$ with random crop along the rescaled long side to $768\times768$. For testing, images in \texttt{07test} and \texttt{minival} are padded to have width and height being the closest multiple of the maximum stride (i.e., $s=64$), to avoid rounding errors. All models were trained for 40 epochs which roughly take 30 and 90 hours for PASCAL VOC \texttt{07+12} and COCO \texttt{trainval35k} respectively.

Following~\cite{Chavali2016,Hosang2016,Lin2014}, we evaluated models at AR with different proposal numbers of 10, 100 and 1,000 $\{\textrm{AR}^{10}, \textrm{AR}^{100}, \textrm{AR}^{1k}\}$ and the area under the curve ($\textrm{AUC}$) of recall across all proposal numbers. Besides, we also evaluated models at AR for different object area $a$: small ($a < 32^{2}$), medium ($32^{2} < a < 96^{2}$) and large ($a > 96^{2}$) with 1,000 proposals $\{\textrm{AR}_{s}^{1k}, \textrm{AR}_{m}^{1k}, \textrm{AR}_{l}^{1k}\}$. It is worth noting that COCO has more complex scenes with diverse and many more small objects than PASCAL VOC does~\cite{Lin2014,Pont2015}. We therefore evaluated all models with PASCAL VOC while selected the top-performance GCN-S and GCN-NS models for COCO evaluation. In the tables, numbers with underline indicate the highest score of a metric among models with different RPN heads, and numbers in bold indicate the highest score of a metric among models with or without using position-sensitivity.

\subsection{Impact of using GCN}
\label{sec:impact-of-using-gcn}
\begin{figure}[t]
	\centering
	\minipage{0.25\textwidth}
	\includegraphics[width=\linewidth]{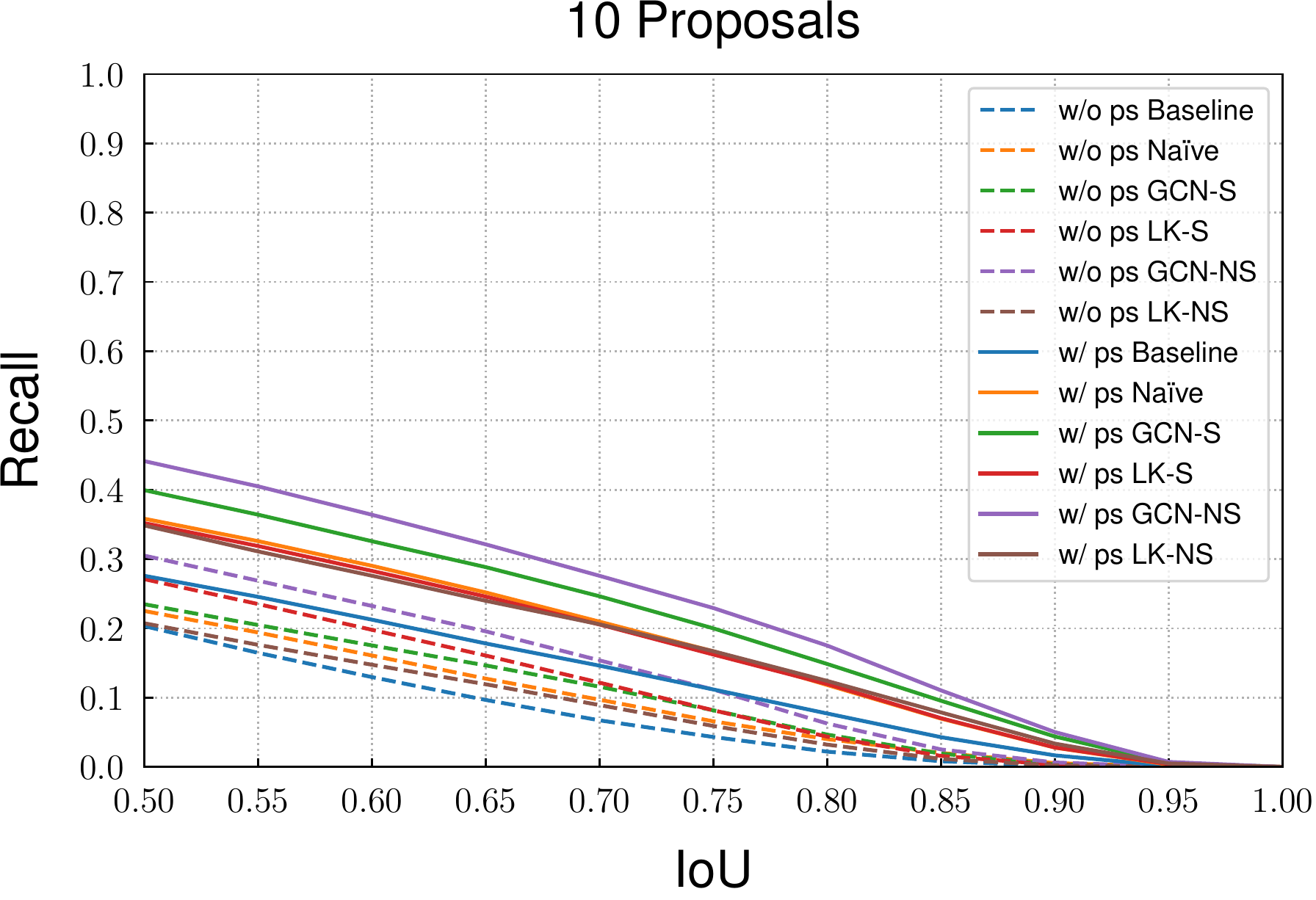}
	\endminipage\hfill
	\minipage{0.25\textwidth}
	\includegraphics[width=\linewidth]{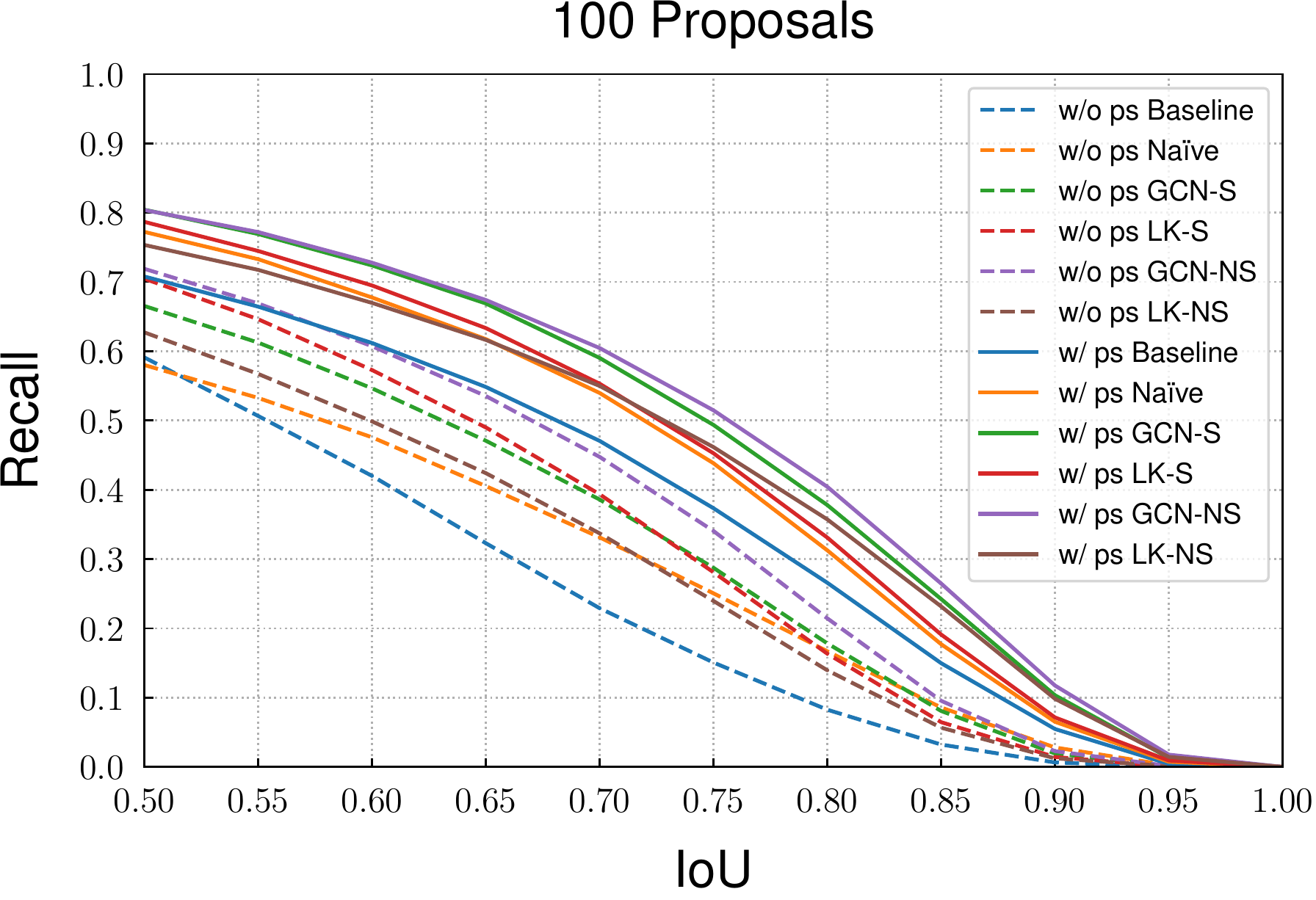}
	\endminipage\hfill
	\minipage{0.25\textwidth}%
	\includegraphics[width=\linewidth]{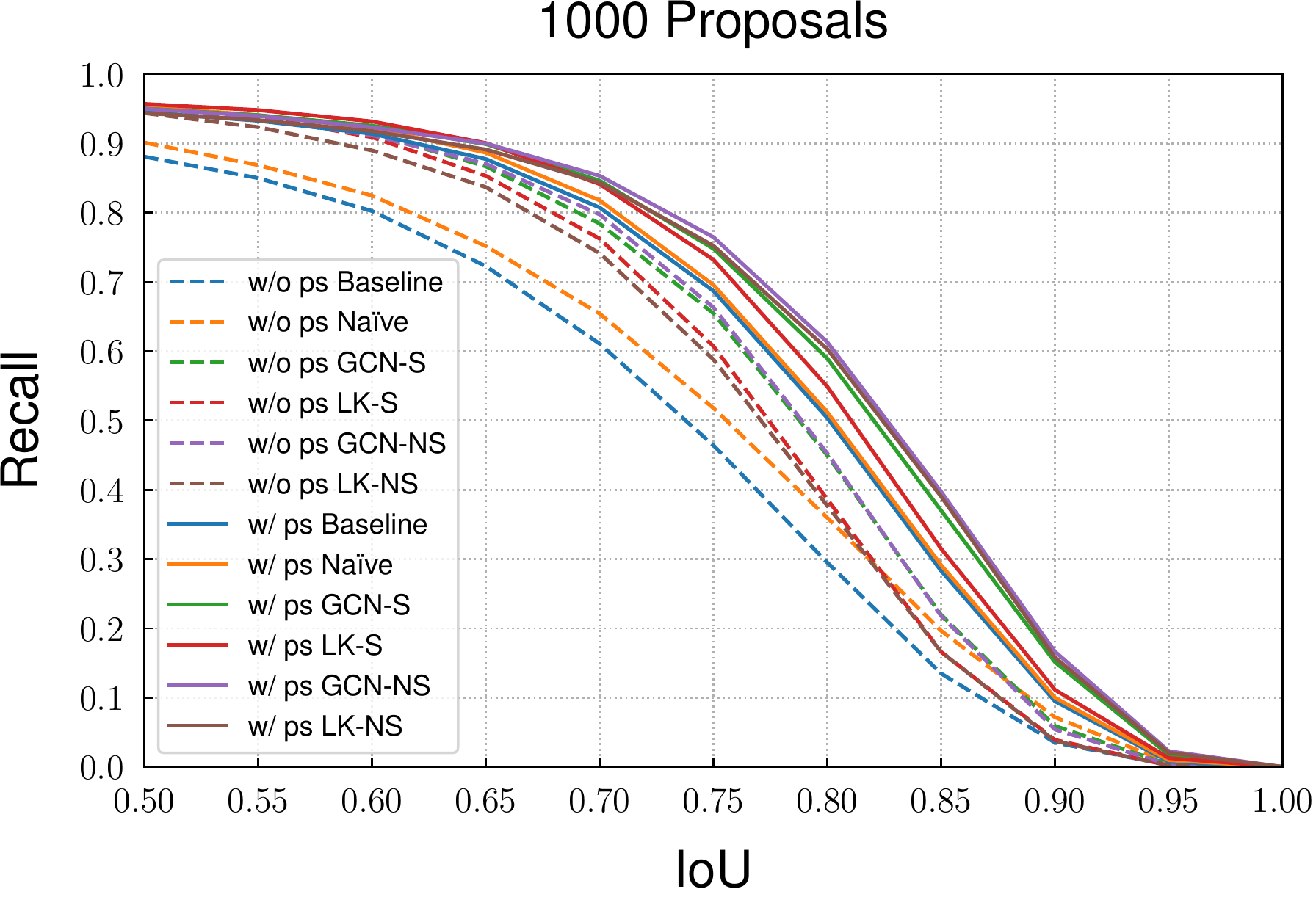}
	\endminipage\hfill
	\minipage{0.25\textwidth}
	\includegraphics[width=\linewidth]{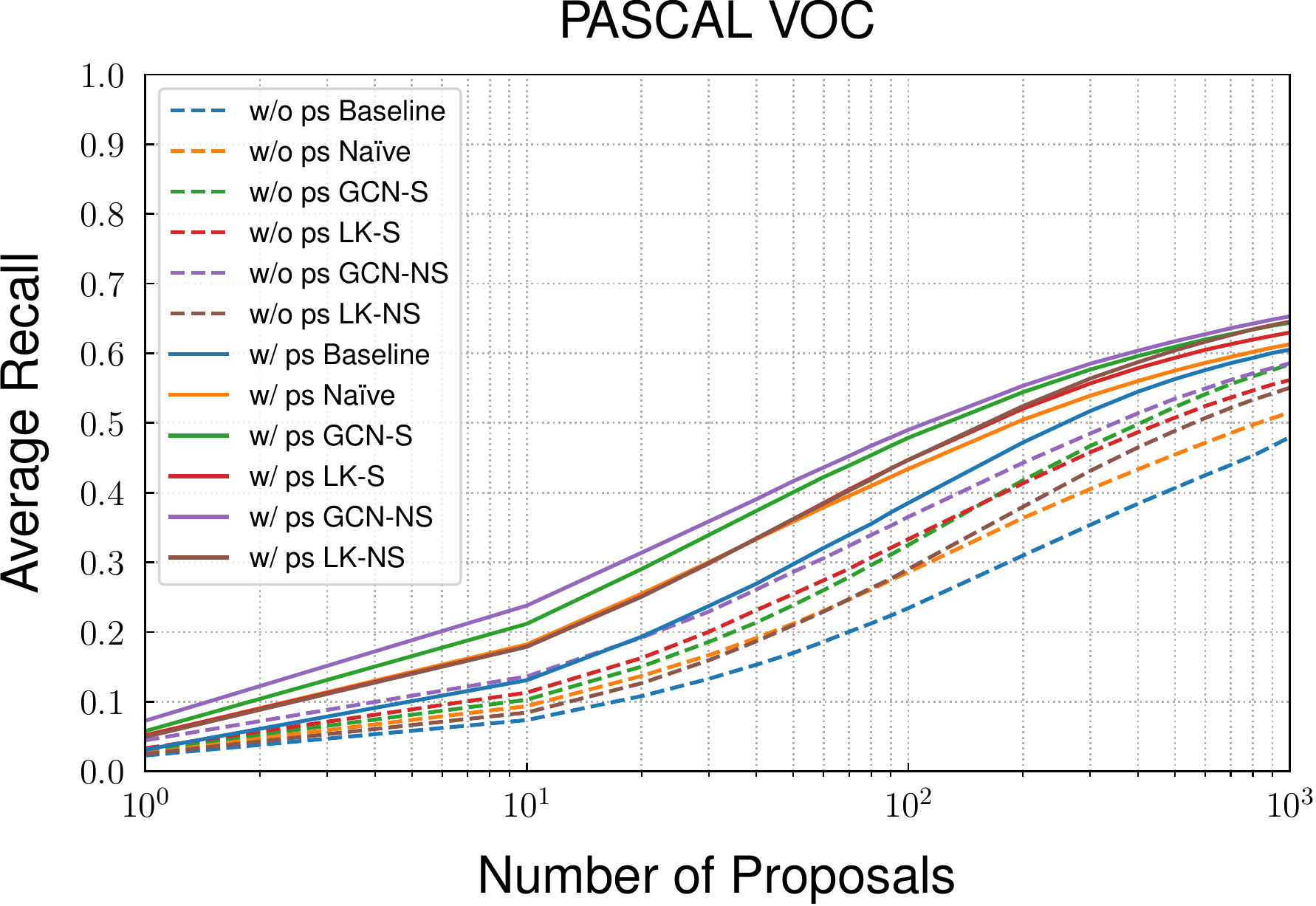}
	\endminipage\hfill
	
	\minipage{0.25\textwidth}
	\includegraphics[width=\linewidth]{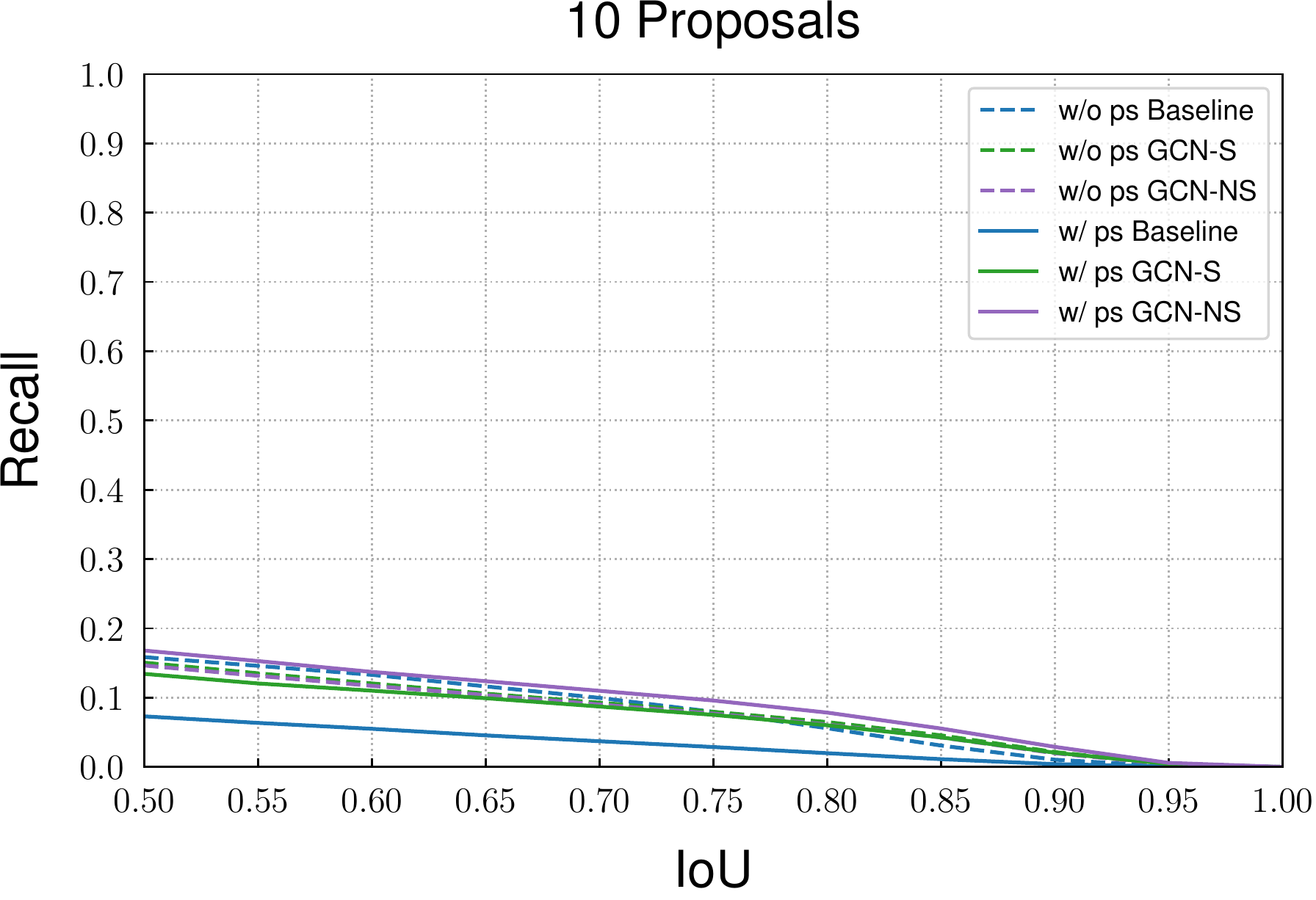}
	\endminipage\hfill
	\minipage{0.25\textwidth}
	\includegraphics[width=\linewidth]{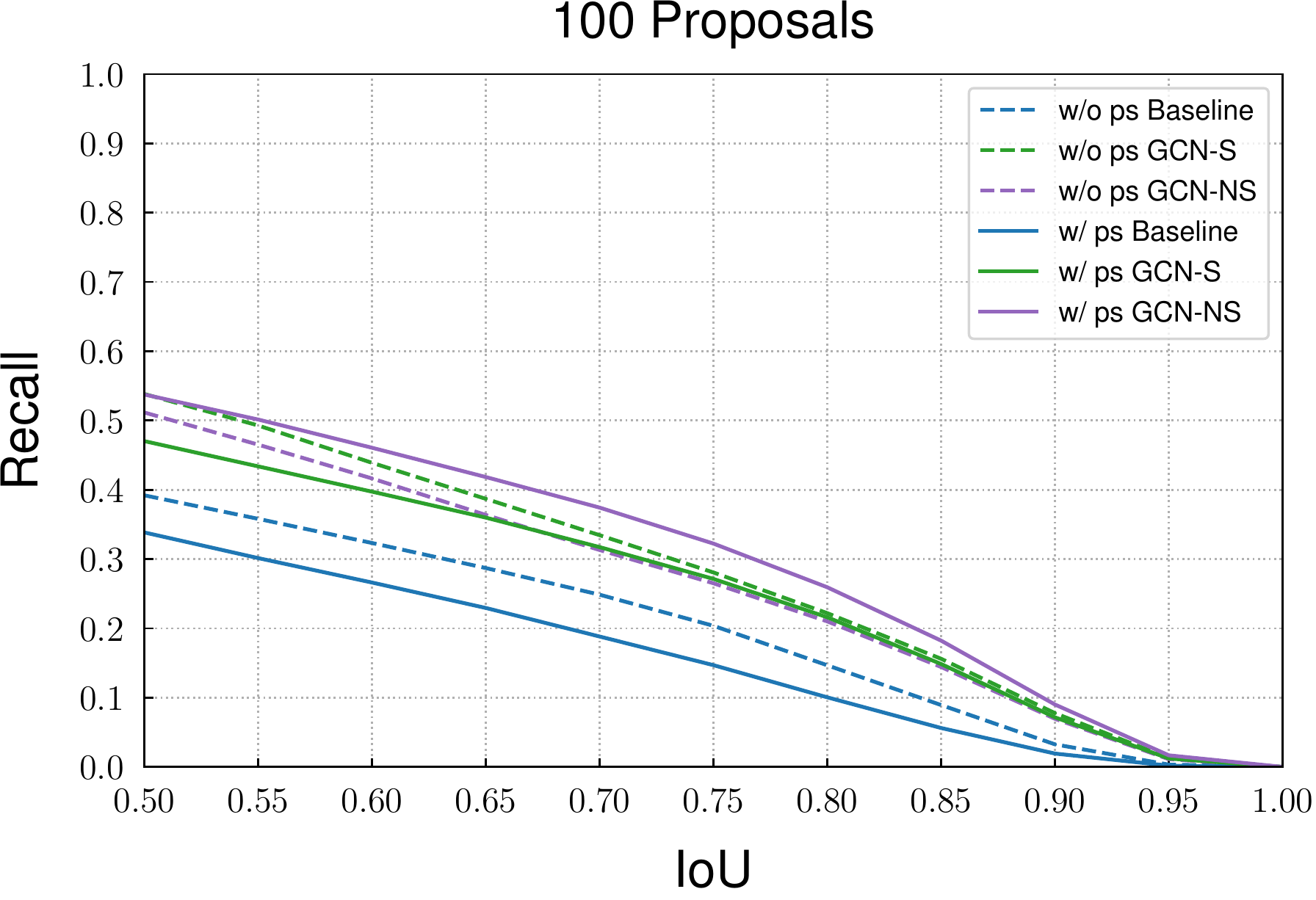}
	\endminipage\hfill
	\minipage{0.25\textwidth}%
	\includegraphics[width=\linewidth]{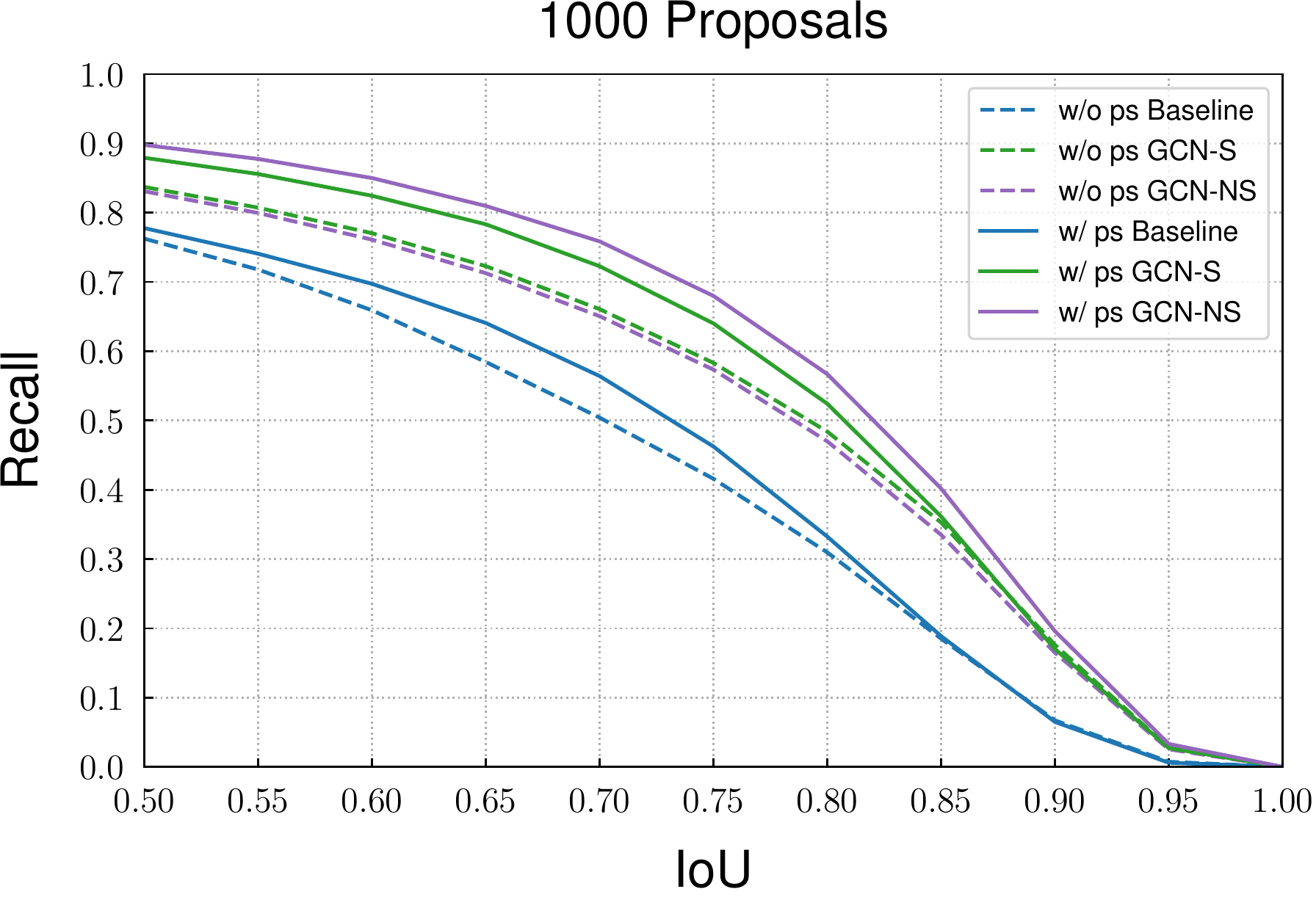}
	\endminipage\hfill
	\minipage{0.25\textwidth}
	\includegraphics[width=\linewidth]{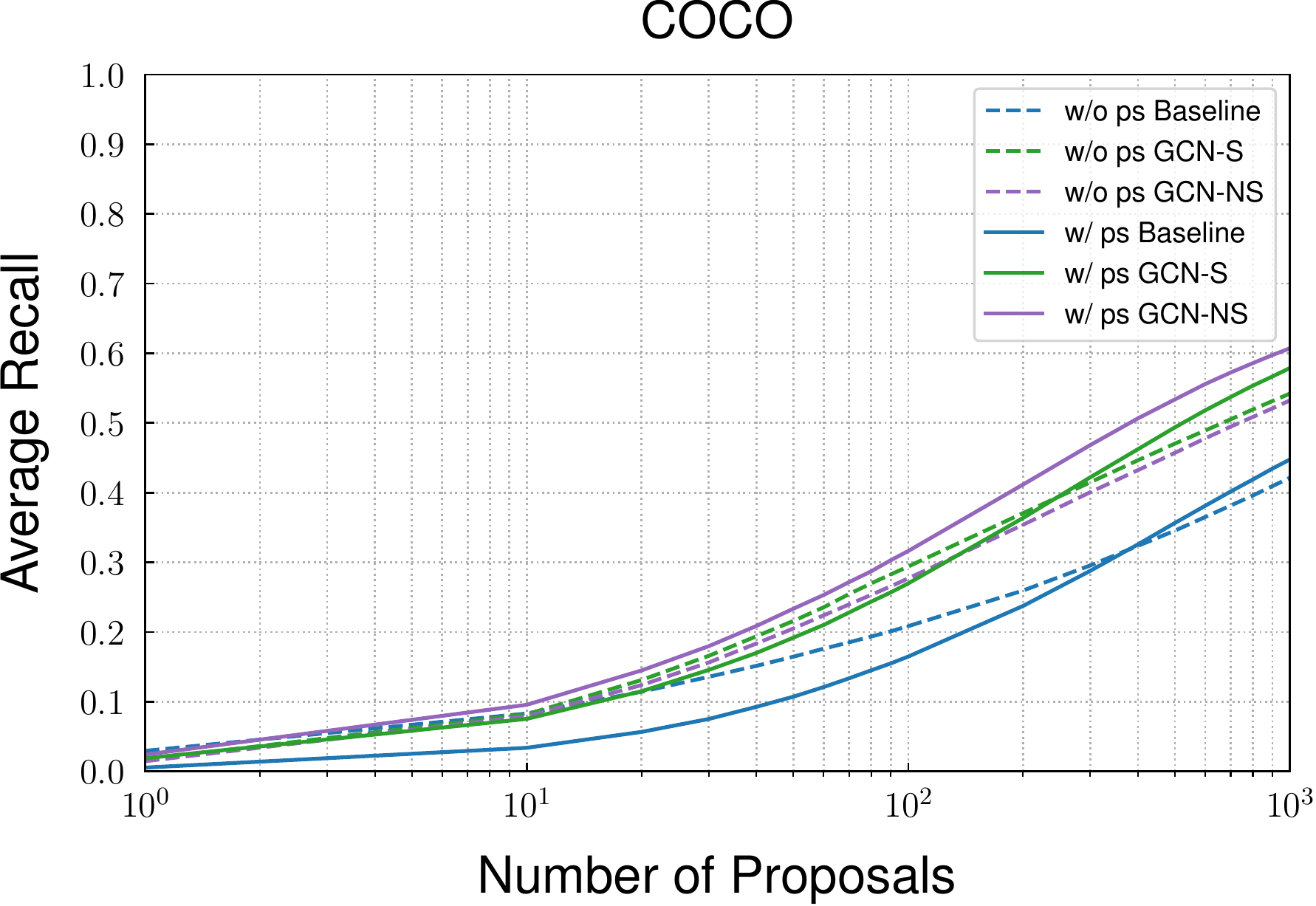}
	\endminipage\hfill

	\minipage{0.25\textwidth}
	\includegraphics[width=\linewidth]{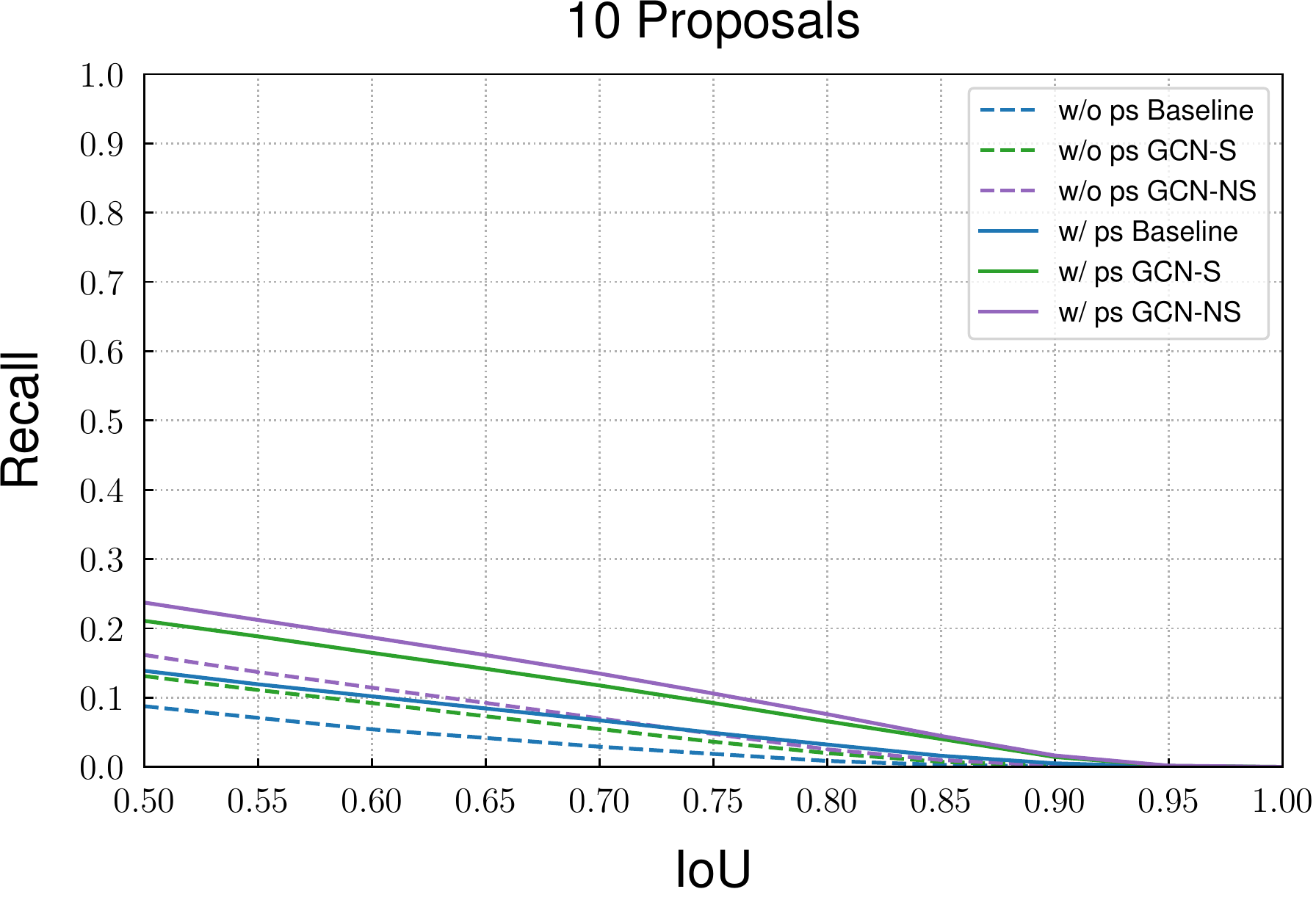}
	\endminipage\hfill
	\minipage{0.25\textwidth}
	\includegraphics[width=\linewidth]{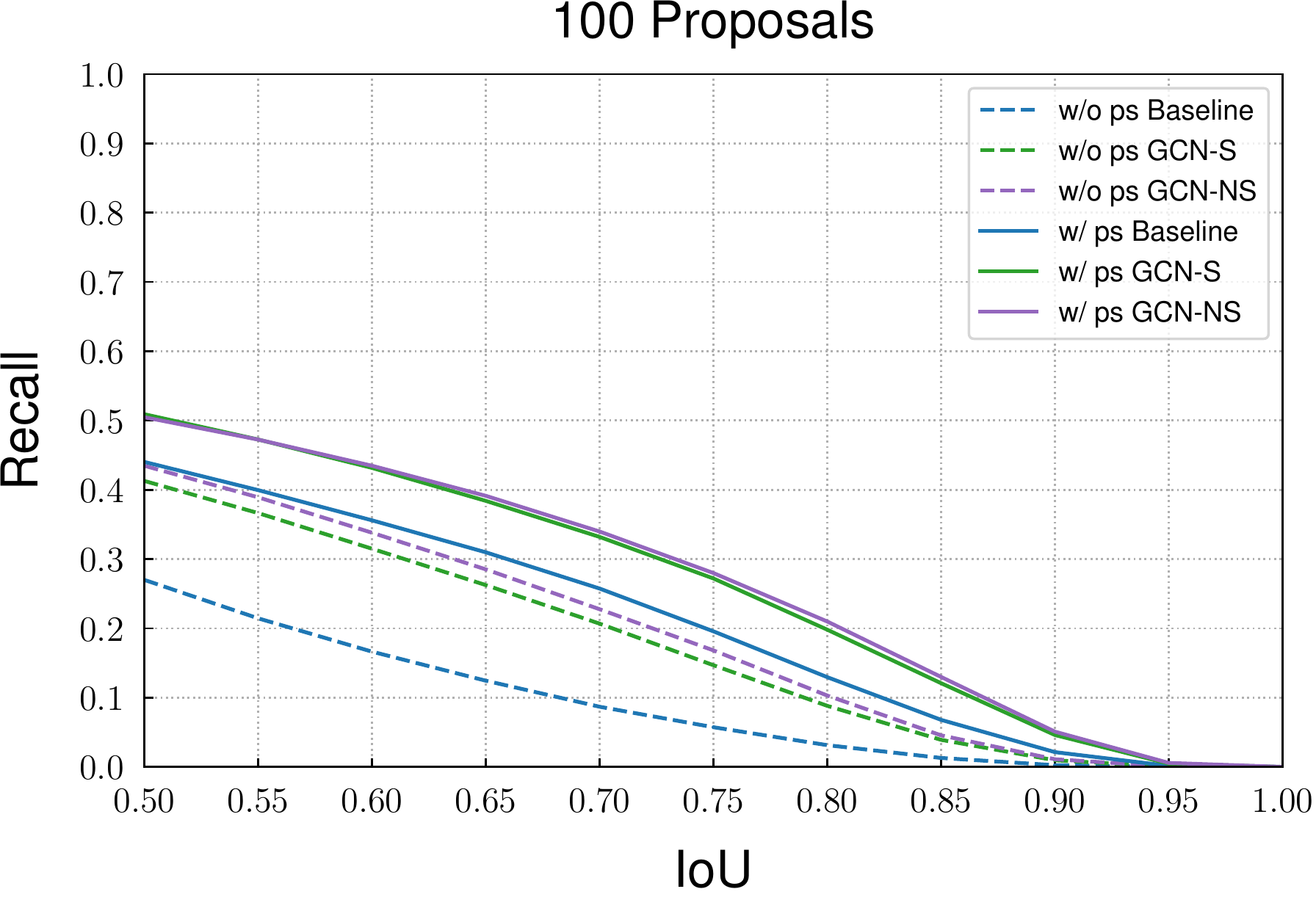}
	\endminipage\hfill
	\minipage{0.25\textwidth}%
	\includegraphics[width=\linewidth]{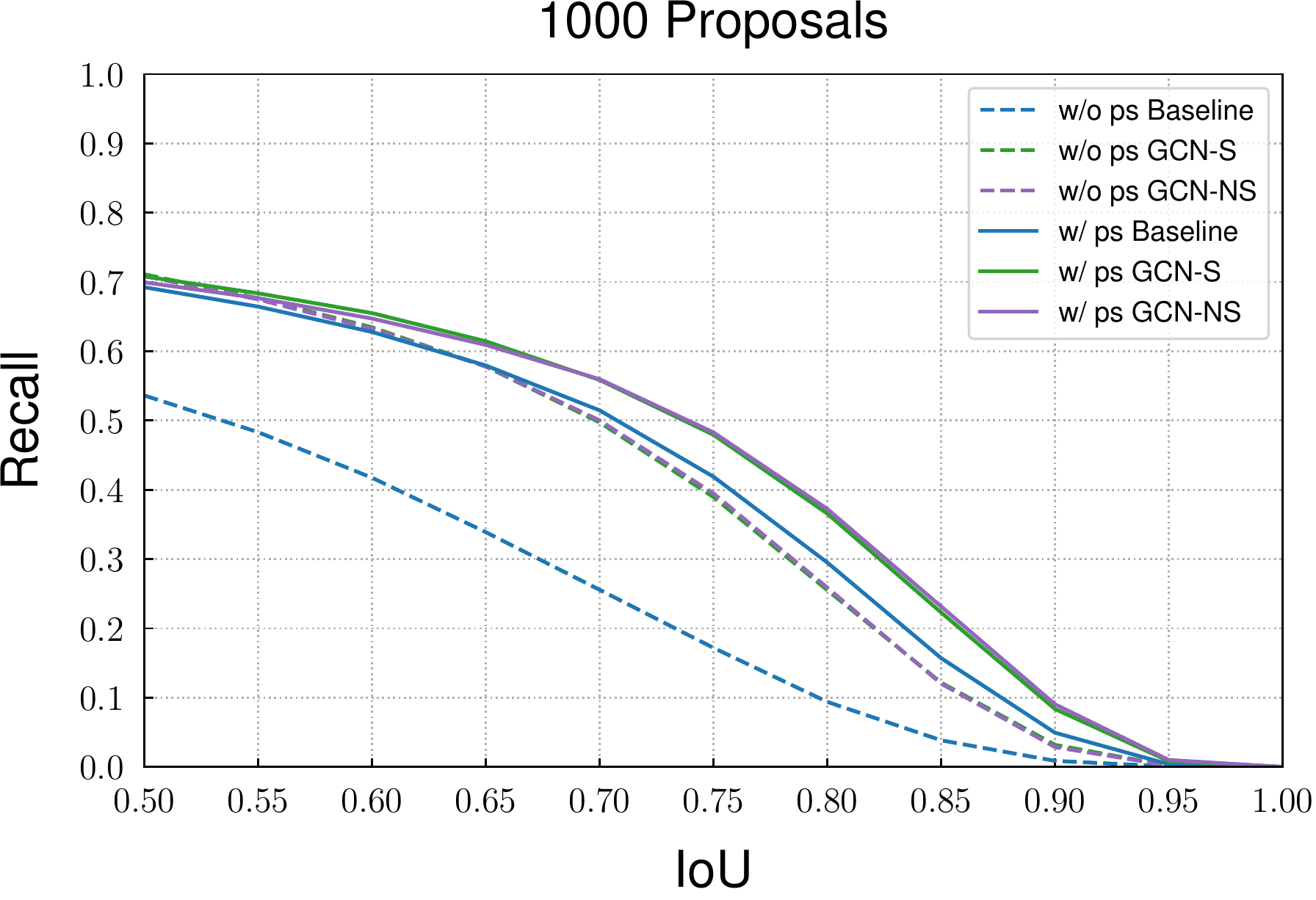}
	\endminipage\hfill
	\minipage{0.25\textwidth}
	\includegraphics[width=\linewidth]{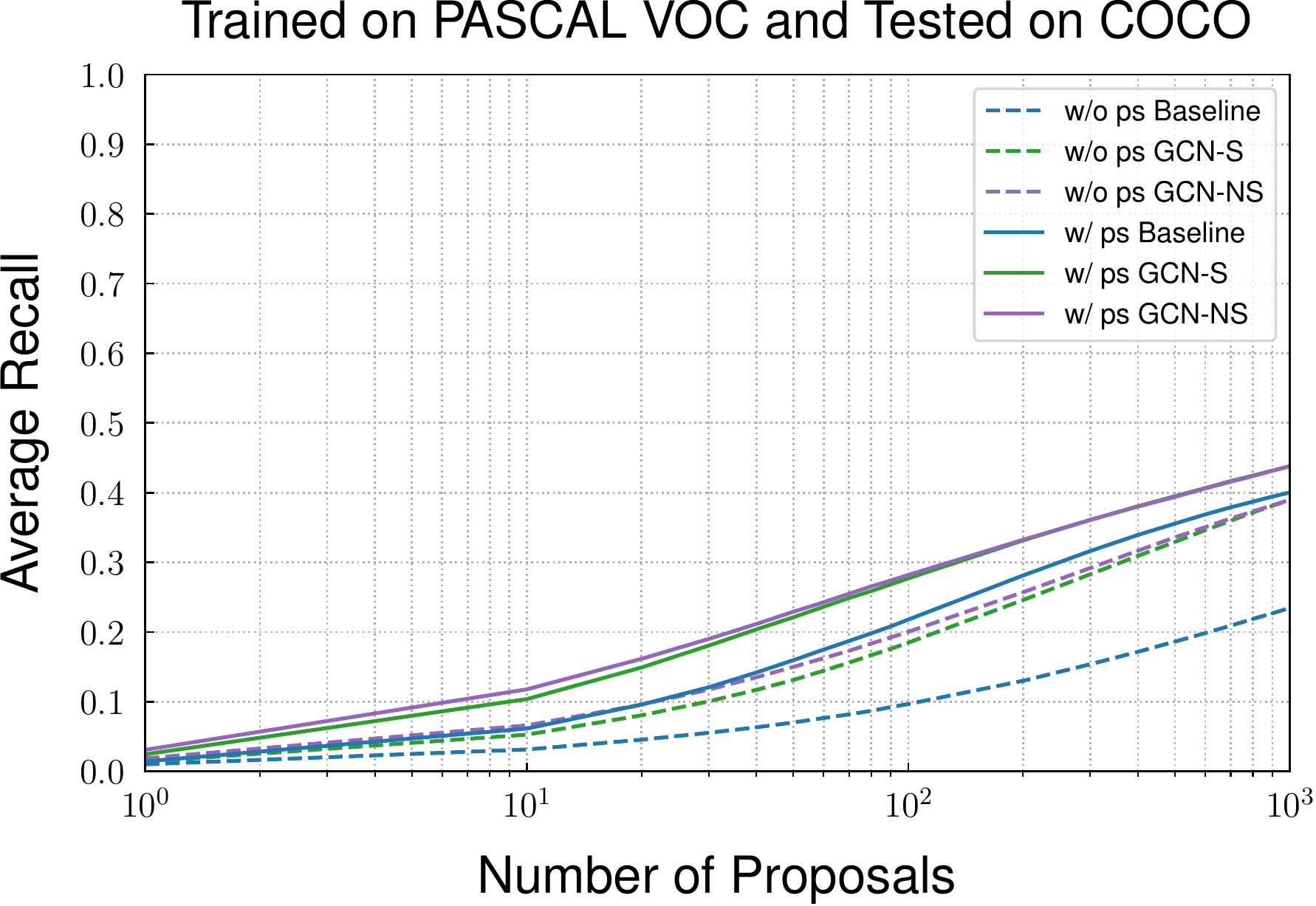}
	\endminipage
	\caption{Recall against IoU with different proposal numbers of 10, 100 and 1,000 and average recall against the number of proposals of all models: The results of PASCAL VOC \texttt{07test} using models trained on PASCAL VOC \texttt{07+12} (\textbf{Row 1}). 
The results of COCO \texttt{minival} using models trained on COCO \texttt{trainval35k} (\textbf{Row 2}) and models trained on PASCAL VOC \texttt{07+12} (\textbf{Row 3}).}
	\label{fig:recall-iou-nprop}
\end{figure}

The results of Table~\ref{tab:pascal-results} and~\ref{tab:coco-results} reveal that by adding GCN in the RPN head, the overall AR can be remarkably improved regardless the number of considered proposals or if the position-sensitivity is employed. This can be also observed in Fig.~\ref{fig:recall-iou-nprop} in which GCN-S and GCN-NS curves are always on the top of others.
\begin{table}[t]
	\caption{Object proposal results of all models trained on PASCAL VOC \texttt{07+12} and evaluated on \texttt{07test}}
	\label{tab:pascal-results}
	\centering
	\resizebox{\textwidth}{!} {\begin{tabular}{lllllllllllllllll}
\toprule
\multicolumn{2}{c}{} & 
\multicolumn{7}{c}{w/o position-sensitive} & \multicolumn{1}{c}{} & 
\multicolumn{7}{c}{w/ position-sensitive} \\
\cmidrule(lr){2-9}
\cmidrule(lr){10-17}
\multicolumn{2}{c}{} &
\multicolumn{1}{c}{$\textrm{AR}^{10}$} &
\multicolumn{1}{c}{$\textrm{AR}^{100}$} &
\multicolumn{1}{c}{$\textrm{AR}^{1k}$} &
\multicolumn{1}{c}{$\textrm{AUC}$} &
\multicolumn{1}{c}{$\textrm{AR}_{s}^{1k}$} &
\multicolumn{1}{c}{$\textrm{AR}_{m}^{1k}$} &
\multicolumn{1}{c}{$\textrm{AR}_{l}^{1k}$} &
\multicolumn{1}{c}{} &
\multicolumn{1}{c}{$\textrm{AR}^{10}$} &
\multicolumn{1}{c}{$\textrm{AR}^{100}$} &
\multicolumn{1}{c}{$\textrm{AR}^{1k}$} &
\multicolumn{1}{c}{$\textrm{AUC}$} &
\multicolumn{1}{c}{$\textrm{AR}_{s}^{1k}$} &
\multicolumn{1}{c}{$\textrm{AR}_{m}^{1k}$} &
\multicolumn{1}{c}{$\textrm{AR}_{l}^{1k}$}
\\
\midrule
Baseline 	& & .074 & .234 & .480 & .272 & .254 & .414 & .566 & & .131 & .385 & .605 & .399 & .423 & .583 & .655 \\ \midrule
Na\"ive 	& & .094 & .286 & .515 & .313 & .410 & .418 & .596 & & .182 & .434 & .613 & .435 & \underline{.466} & .593 & .655 \\
GCN-S 		& & .103 & .325 & .584 & .356 & \underline{\textbf{.471}} & .558 & .623 & & .212 & .479 & .644 & .471 & .445 & .603 & .709 \\
LK-S 		& & .113 & .333 & .562 & .356 & .441 & .547 & .595 & & .178 & .447 & .630 & .446 & .463 & \underline{\textbf{.613}} & .674 \\
GCN-NS 		& & \underline{.136} & \underline{.365} & \underline{.586} & \underline{.383} & .445 & \underline{.569} & \underline{.625} & & \underline{\textbf{.238}} & \underline{\textbf{.490}} & \underline{\textbf{.653}} & \underline{\textbf{.484}} & .453 & .593 & \textbf{\underline{.730}} \\
LK-NS 		& & .084 & .290 & .551 & .326 & .420 & .553 & .575 & & .179 & .447 & .645 & .450 & .429 & .582 & .728 \\
\bottomrule
\end{tabular}%
}
\end{table}
\begin{table}[t]
	\caption{Object proposal results of all models trained on COCO \texttt{trainval35k} and evaluated on \texttt{minival}}
	\label{tab:coco-results}
	\centering
	\resizebox{\textwidth}{!} {\begin{tabular}{lllllllllllllllll}
\toprule
\multicolumn{2}{c}{} & 
\multicolumn{7}{c}{w/o position-sensitive} & \multicolumn{1}{c}{} & 
\multicolumn{7}{c}{w/ position-sensitive} \\
\cmidrule(lr){2-9}
\cmidrule(lr){10-17}
\multicolumn{2}{c}{} &
\multicolumn{1}{c}{$\textrm{AR}^{10}$} &
\multicolumn{1}{c}{$\textrm{AR}^{100}$} &
\multicolumn{1}{c}{$\textrm{AR}^{1k}$} &
\multicolumn{1}{c}{$\textrm{AUC}$} &
\multicolumn{1}{c}{$\textrm{AR}_{s}^{1k}$} &
\multicolumn{1}{c}{$\textrm{AR}_{m}^{1k}$} &
\multicolumn{1}{c}{$\textrm{AR}_{l}^{1k}$} &
\multicolumn{1}{c}{} &
\multicolumn{1}{c}{$\textrm{AR}^{10}$} &
\multicolumn{1}{c}{$\textrm{AR}^{100}$} &
\multicolumn{1}{c}{$\textrm{AR}^{1k}$} &
\multicolumn{1}{c}{$\textrm{AUC}$} &
\multicolumn{1}{c}{$\textrm{AR}_{s}^{1k}$} &
\multicolumn{1}{c}{$\textrm{AR}_{m}^{1k}$} &
\multicolumn{1}{c}{$\textrm{AR}_{l}^{1k}$}
\\
\midrule
Baseline 	& & \underline{.083} & .208 & .421 & .242 & .308 & \underline{.562} & .392 & & .034 & .165 & .448 & .219 & .385 & .411 & .566 \\ \midrule
GCN-S		& & .082 & \underline{.294} & \underline{.542} & \underline{.322} & .414 & .558 & \underline{.680} & & .075 & .270 & .579 & .321 & .485 & .592 & .677 \\
GCN-NS 		& & .079 & .277 & .532 & .310 & \underline{.422} & .552 & .643 & & \underline{\textbf{.096}} & \underline{\textbf{.316}} & \underline{\textbf{.607}} & \underline{\textbf{.358}} & \underline{\textbf{.493}} & \underline{\textbf{.623}} & \underline{\textbf{.726}} \\
\bottomrule
\end{tabular}%
}
\end{table}

$\textrm{AR}^{1k}_{s}$ in particular benefits from learning the global image context. One can observe that compared to \textit{Baseline} model, the scores have been boosted by 85\% from 0.254 to 0.471 on PASCAL VOC with GCN-S model, and by 37\% from 0.308 to 0.422 on COCO with GCN-NS. Therefore, the $\textrm{AR}^{1k}$ has been overall improved from 0.480 and 0.421 to 0.586 and 0.542, which are 22\% and 29\% respectively. Fig.~\ref{fig:recall-iou-nprop} also shows that GCN-S and GCN-NS models have the highest recall scores across all IoU thresholds with different proposal numbers. 

One may argue that the improvement in GCN-S and GCN-NS models is due to the increased number of parameters. From Table~\ref{tab:pascal-results}, LK-S and LK-NS models have also shown some improvement but considering the extra number of parameters compared with \textit{Baseline} model, they are not as effective as GCN-S and GCN-NS models. This shows that using separable convolution kernel matters, which aligns with the observation in~\cite{Peng2017}. \textit{Na\"ive} model also exhibits similar results.

\subsection{Impact of using position-sensitivity}
\label{sec:impact-of-using-position-sensitivity}
\begin{figure}[t]
	\centering
	\includegraphics[width=\textwidth]{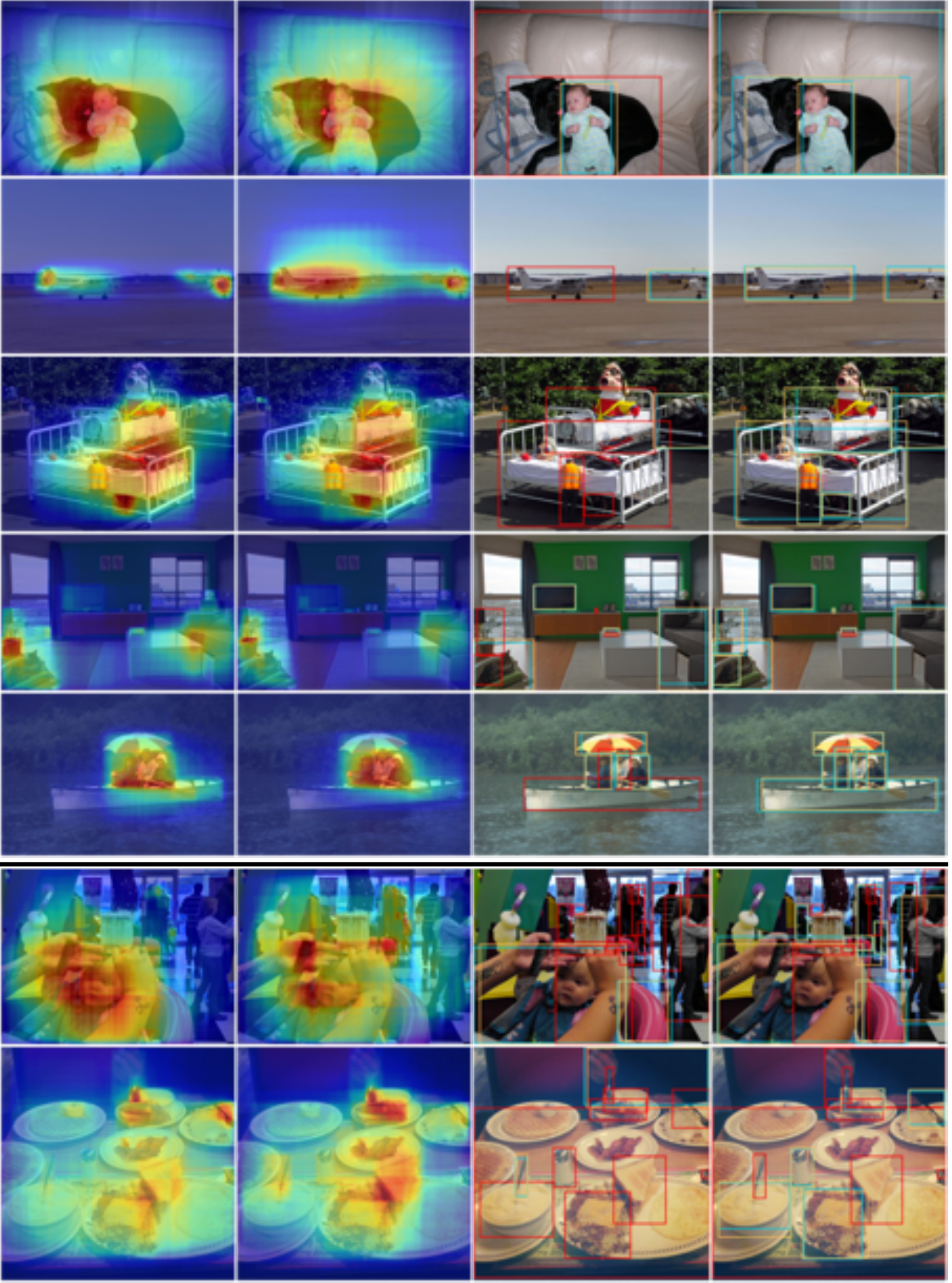}
	\caption{The impact of position-sensitivity: visualisation on the distribution heatmap and hit-and-miss of the top 1,000 proposals by GCN-NS models for a number of selected PASCAL VOC \texttt{07test} (\textbf{Row 1-2}) and COCO \texttt{minival} (\textbf{Row 3-5}) images. For each pair of images, model without position-sensitivity is on the left and the one with position-sensitivity is on the right. \textbf{Col 1-2:} The heatmaps are plotted by stacking the proposal boxes. \textbf{Col 3-4:} The bounding boxes in orange are ground truth boxes with their corresponding hit proposals in cyan, in which the IoU threshold is set to 0.7, and the bounding boxes in red are missed cases. \textbf{Row 6-7:} All models tend to fail in images with complex scenes and diverse object aspect ratios. Nonetheless, note that models with position-sensitivity generally have higher hit rate. Sec.~\ref{sec:impact-of-using-position-sensitivity} for detailed discussions.}
	\label{fig:hit-miss-exp}
\end{figure}

As shown in Table~\ref{tab:pascal-results}, on PASCAL VOC, among different proposal numbers and object sizes, models using position-sensitive components generally result in higher AR. Specifically, for $\textrm{AR}^{1k}$, \textit{Baseline} model shows an improvement from 0.480 to 0.605 (26\%) and GCN-NS from 0.586 to 0.653 (11\%). As shown in Table~\ref{tab:coco-results}, the experiment on COCO shows similar results, in which \textit{Baseline} model has an improvement from 0.421 to 0.448 (6\%) and GCN-NS model from 0.532 to 0.607 (14\%). 
To investigate on the small object proposals, $\textrm{AR}_{s}^{1k}$ reveal that training on a large number of annotated small objects in COCO indeed helps in higher $\textrm{AR}_{s}^{1k}$ scores, compared with the results of PASCAL VOC counterpart. GCN-NS has achieved much higher $\textrm{AR}_{s}^{1k}$ ($0.493$) score, which is a $17\%$ improvement compared to the counterpart. 
Fig.~\ref{fig:hit-miss-exp} visualises the distribution heatmap and hit-and-miss of top 1,000 proposals using GCN-NS models with and without taking position-sensitivity into account, in which the hits are with a threshold 0.7 for the ground truth IoU. One can qualitatively tell that by using the position-sensitivity, models can generate proposals closer to objects and thus result in more hits, especially for objects with extremely large or small sizes. 

\subsection{Inference time}
\label{sec:inference-time}
Introducing both GCN structure and position-sensitive score maps in the RPN head brings in more learnable parameters resulting in more computation.
Beside the input image size which has a directly impact on the overall inference time, the number of anchors, the kernel size of GCN and the grid number $k$ of position-sensitive score maps are also key factors. Table~\ref{tab:model-param-time} lists the models' inference times averaged on all input images in both \texttt{07test} and \texttt{minival}, in which \textit{Baseline} model shows a performance of 26.6/58.2 (denoted for \texttt{07test}/\texttt{minival}) ms. 
Adding the position-sensitive score maps (with grid size $k=4$) takes extra 9.1/21.3 ms, 9.8/19.8 ms and 8.9/20.1 ms for \textit{Baseline}, GCN-S and GCN-NS model respectively, which show a comparable time difference. In contrast, introducing the GCN structure in the \textit{Baseline} model adds 7.7/18.1 ms for GCN-S model while additional 9.3/25.3 ms for GCN-NS model. This reveals that using non-shared smoother generally takes more times than using shared smoother does. GCN-NS with position-sensitivity, as the best performance model, has a running time 44.8 ms ($\sim22$ fps) for \texttt{07+12} and 103.6 ms ($\sim10$ fps) for \texttt{minival}.

\subsection{Model generalisation}
\label{sec:model-generalisation}
\begin{figure}[t]
	\centering
	\includegraphics[width=\textwidth]{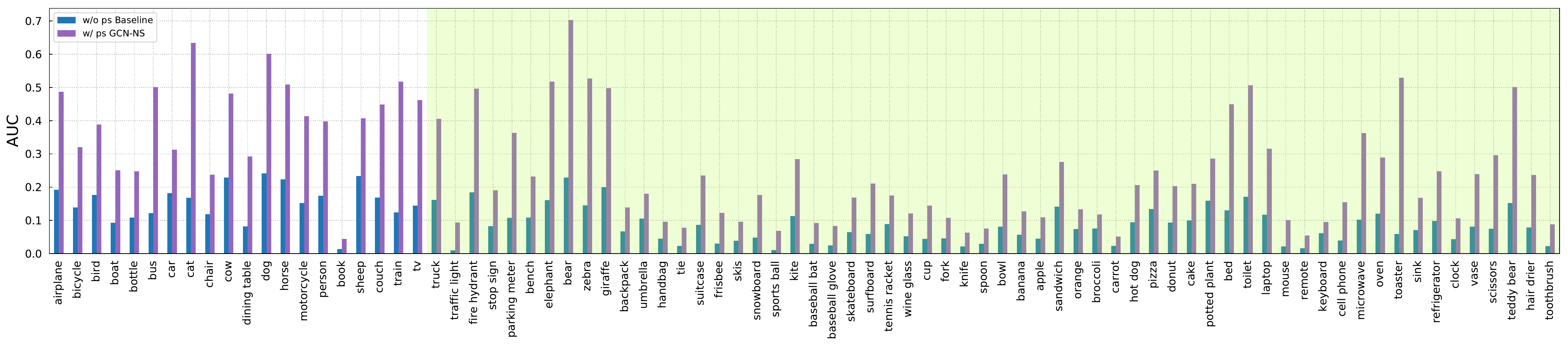}
	\caption{Per-category AUC performance of models trained on PASCAL VOC \texttt{07+12} and evaluated on COCO \texttt{minival}. Common and non-common categories are split in the white and green region respectively.}
	\label{fig:common-non-common}
\end{figure}

\begin{figure}[t]
	\centering
	\includegraphics[width=\textwidth]{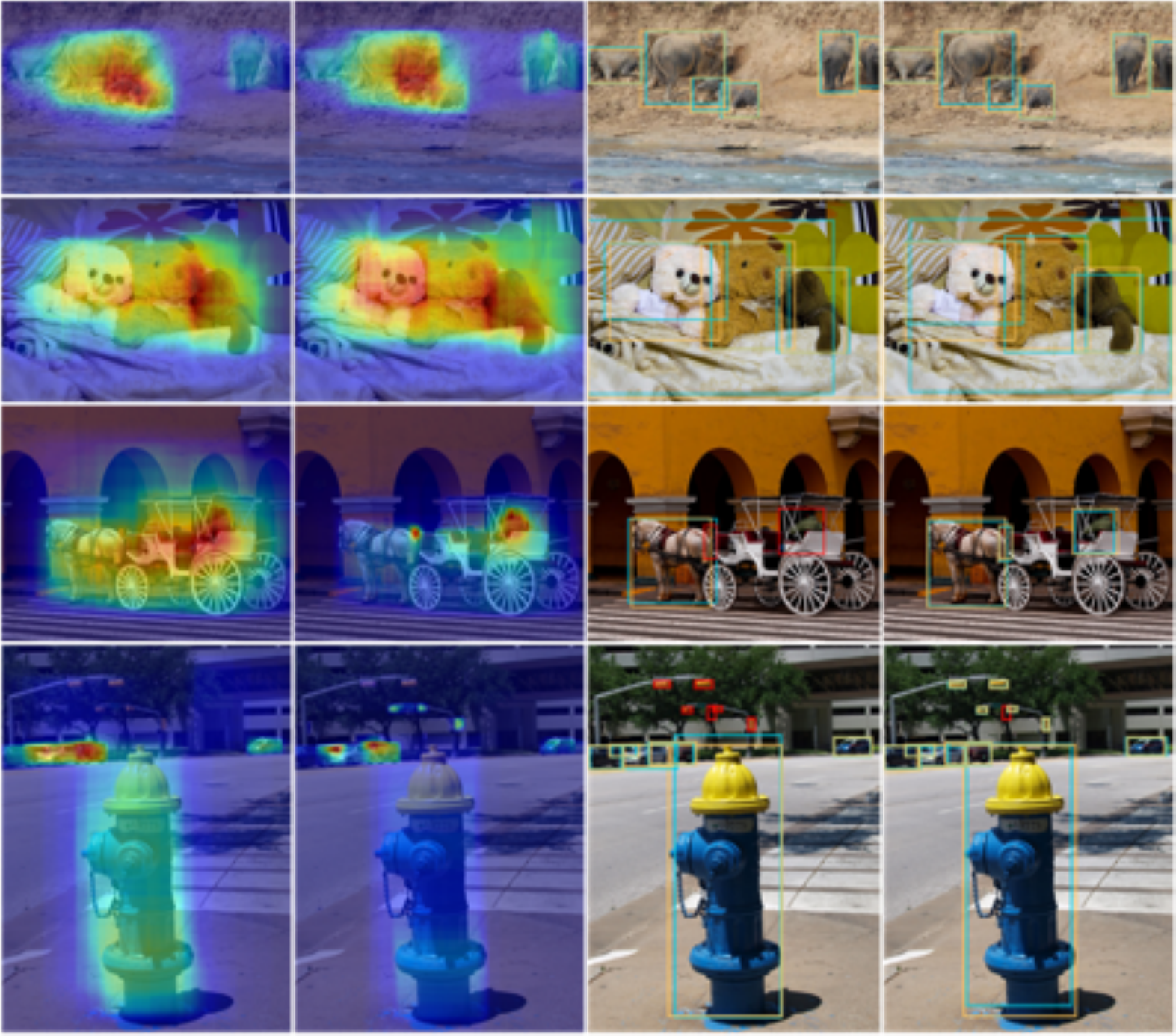}
	\caption{The results of model generalisation experiments visualised in the distribution heatmap and hit-and-miss of the top 1,000 proposals by GCN-NS models for a number of selected COCO \texttt{minival}. For each pair of COCO \texttt{minival} images, result of model trained on PASCAL VOC is on the left and the one of model trained on COCO is on the right. See Sec.~\ref{sec:model-generalisation} for detailed discussions.}
	\label{fig:generalisation-hit-miss-exp}
\end{figure}

To evaluate the generalisation ability of the proposed method, the models trained on PASCAL VOC \texttt{07+12} are used to evaluate on COCO \texttt{minival}. Note that compared to COCO (80 categories), PASCAL VOC is a much smaller dataset with limited object categories (20 categories) and almost all categories in PASCAL VOC are included in COCO.  We separate the categories of COCO into two sets: \textit{common} and \textit{non-common}. Common categories are ones in PASCAL VOC, while non-common are unseen categories. In addition, image scenes in PASCAL VOC are relatively simple with less small objects annotations. It is therefore more challenging for a model learned from PASCAL VOC to performance object proposal on COCO images. 
The results are shown in Table~\ref{tab:pascal-coco-results} and Fig.~\ref{fig:recall-iou-nprop} (\textbf{Row 3}). Surprisingly, the $\textrm{AR}^{1k}$ of the best performance models, GCN-S and GCN-NS with position-sensitivity, trained with PASCAL VOC \texttt{07+12} can still outperform \textit{Baseline} model trained with COCO \texttt{trainval35k} (i.e., 0.438 vs. 0.421). It is expected that the model will not work well on small objects since PASCAL VOC does not contain many small objects, but nevertheless the model still shows decent performance on large objects in which the $\textrm{AR}^{1k}_{l}$ is up to 0.693 as shown in Table~\ref{tab:pascal-coco-results}. The score is comparable to the other models trained on COCO \texttt{trainval35k} as shown in Table~\ref{tab:coco-results}.
Delving into the details, we show the breakdown results of the model generalisation experiment for common and non-common categories are shown in Table \ref{tab:pascal-coco-comvsnoncom-results}. All models have better performance on common categories overall. However, compared to the \textit{Baseline} model, the proposed components significantly improved the performance on both common and non-common categories. The per-category AUC performance in Fig.~\ref{fig:common-non-common} shows that non-common categories do not necessarily have worse performance than common categories (e.g., bear, zebra, and toilet). This indicates that the proposed object proposal networks are able to generalise proposals from a smaller to a larger and more complex dataset, from common to non-common categories, and that the proposed architecture can further improve the generalisation for all categories.
Fig.~\ref{fig:generalisation-hit-miss-exp} qualitatively demonstrates the generalisation ability of the proposed method. 
Although the model trained on PASCAL VOC \texttt{07+12} fails at detecting small objectness, it still exhibits a certain degree of generalisation to unseen categories (e.g., elephant, teddy bear, or fire hydrant).
\begin{table}
	\caption{Object proposal results of all models trained on PASCAL VOC \texttt{07+12} and evaluated on COCO \texttt{minival}}
	\label{tab:pascal-coco-results}
	\centering
	\resizebox{\textwidth}{!} {\begin{tabular}{lllllllllllllllll}
\toprule
\multicolumn{2}{c}{} & 
\multicolumn{7}{c}{w/o position-sensitive} & \multicolumn{1}{c}{} & 
\multicolumn{7}{c}{w/ position-sensitive} \\
\cmidrule(lr){2-9}
\cmidrule(lr){10-17}
\multicolumn{2}{c}{} &
\multicolumn{1}{c}{$\textrm{AR}^{10}$} &
\multicolumn{1}{c}{$\textrm{AR}^{100}$} &
\multicolumn{1}{c}{$\textrm{AR}^{1k}$} &
\multicolumn{1}{c}{$\textrm{AUC}$} &
\multicolumn{1}{c}{$\textrm{AR}_{s}^{1k}$} &
\multicolumn{1}{c}{$\textrm{AR}_{m}^{1k}$} &
\multicolumn{1}{c}{$\textrm{AR}_{l}^{1k}$} &
\multicolumn{1}{c}{} &
\multicolumn{1}{c}{$\textrm{AR}^{10}$} &
\multicolumn{1}{c}{$\textrm{AR}^{100}$} &
\multicolumn{1}{c}{$\textrm{AR}^{1k}$} &
\multicolumn{1}{c}{$\textrm{AUC}$} &
\multicolumn{1}{c}{$\textrm{AR}_{s}^{1k}$} &
\multicolumn{1}{c}{$\textrm{AR}_{m}^{1k}$} &
\multicolumn{1}{c}{$\textrm{AR}_{l}^{1k}$}
\\
\midrule
Baseline 	& & .031 & .097 & .234 & .122 & .114 & .234 & .382 & & .061 & .218 & .400 & .240 & .224 & .401 & .614 \\ \midrule
GCN-S		& & .053 & .185 & \underline{.390} & .217 & \underline{\textbf{.240}} & \underline{.430} & .524 & & .104 & .277 & \underline{\textbf{.438}} & .288 & .227 & \underline{\textbf{.463}} & .665 \\
GCN-NS 		& & \underline{.066} & \underline{.200} & \underline{.390} & \underline{.228} & .239 & .420 & \underline{.538} & & \underline{\textbf{.118}} & \underline{\textbf{.282}} & \underline{\textbf{.438}} & \underline{\textbf{.292}} & \underline{.235} & .432 & \underline{\textbf{.693}} \\
\bottomrule
\end{tabular}%
}
\end{table}
\begin{table}
	\begin{minipage}[t]{\linewidth} 
		\caption{Object proposal results of all models trained on PASCAL VOC \texttt{07+12} and evaluated on COCO \texttt{minival} for \textit{common} and \textit{non-common} categories. Note that \textit(non) denotes models evaluated on \textit{non-common} categories}
		\label{tab:pascal-coco-comvsnoncom-results}
		\resizebox{\textwidth}{!} {\begin{tabular}{lllllllllllllllll}
	\toprule
	\multicolumn{2}{c}{} & 
	\multicolumn{7}{c}{w/o position-sensitive} & \multicolumn{1}{c}{} & 
	\multicolumn{7}{c}{w/ position-sensitive} \\
	\cmidrule(lr){2-9}
	\cmidrule(lr){10-17}
	\multicolumn{2}{c}{} &
	\multicolumn{1}{c}{$\textrm{AR}^{10}$} &
	\multicolumn{1}{c}{$\textrm{AR}^{100}$} &
	\multicolumn{1}{c}{$\textrm{AR}^{1k}$} &
	\multicolumn{1}{c}{$\textrm{AUC}$} &
	\multicolumn{1}{c}{$\textrm{AR}_{s}^{1k}$} &
	\multicolumn{1}{c}{$\textrm{AR}_{m}^{1k}$} &
	\multicolumn{1}{c}{$\textrm{AR}_{l}^{1k}$} &
	\multicolumn{1}{c}{} &
	\multicolumn{1}{c}{$\textrm{AR}^{10}$} &
	\multicolumn{1}{c}{$\textrm{AR}^{100}$} &
	\multicolumn{1}{c}{$\textrm{AR}^{1k}$} &
	\multicolumn{1}{c}{$\textrm{AUC}$} &
	\multicolumn{1}{c}{$\textrm{AR}_{s}^{1k}$} &
	\multicolumn{1}{c}{$\textrm{AR}_{m}^{1k}$} &
	\multicolumn{1}{c}{$\textrm{AR}_{l}^{1k}$}
	\\
	\midrule
	Baseline & & .046 & .129 & .285 & .156 & .180 & .316 & .369 & & .076 & .271 & .483 & .294 & .329 & .506 & .629 \\ 
	Baseline (\textit{non}) & & .013 & .055 & .170 & .079 & .035 & .139 & .402 & & .043 & .151 & .295 & .171 & .099 & .280 & .590 \\ \midrule
	GCN-S & & .061 & .217 & .447 & .252 & \underline{.329} & \underline{.499} & .524 & & .128 & .344 & .520 & .351 & .327 & \underline{\textbf{.564}} & .687 \\
	GCN-S (\textit{non}) & & .042 & .144 & .317 & .172 & .134 &  .349 &  .525 & & .073 &  .192 &  .334 &  .207 & .107 &  .346 & .632 \\
	GCN-NS & & \underline{.078} & \underline{.240} & \underline{.451} & \underline{.269} & .323 & .493 & \underline{.549} & & \underline{\textbf{.149}} & \underline{\textbf{.353}} & \underline{\textbf{.524}} & \underline{\textbf{.360}} & \underline{\textbf{.335}} & .540 & \underline{\textbf{.716}} \\
	GCN-NS (\textit{non}) 		& &  .050 &  .150 &  .312 &  .175 &  .138 & .334 & .522 & &  .078 &  .192 & .328 & .206 &  .114 & .305 &  .658 \\
	\bottomrule
\end{tabular}%
	}
	\end{minipage}%
\end{table}

\section{Conclusions}
In this paper, we have proposed object proposal networks based on the observation that accurate detection relies on translation-invariance for objectness classification, translation-variance for localisation as regression and scale-invariance for various object sizes. Thorough experiments on PASCAL VOC and COCO datasets have shown that the adoption of global convolutional network (GCN) and position-sensitivity components can significantly improve object proposal performance while keeping the network lightweight to achieve real-time performance.

%
%
\clearpage
\bibliographystyle{splncs04}

\beginsupplement

\title{Toward Scale-Invariance and Position-Sensitive Object Proposal Networks\\ 
	Supplementary Material} 

\titlerunning{Toward Scale-Inv. and Pos.-Sens. Reg. Propos. Net.}

\author{Hsueh-Fu Lu \and
	Xiaofei Du \and
	Ping-Lin Chang}

\authorrunning{H.F. Lu, X. Du  and P.L. Chang}

\institute{Umbo Computer Vision Inc. \\
	\email{\{topper.lu, xiaofei.du, ping-lin.chang\}@umbocv.com}\\ 
	\url{https://umbocv.ai}
}

\maketitle

\section{Overview}
\begin{figure}[t]
	\centering
	\minipage{0.25\textwidth}
	\includegraphics[width=\linewidth]{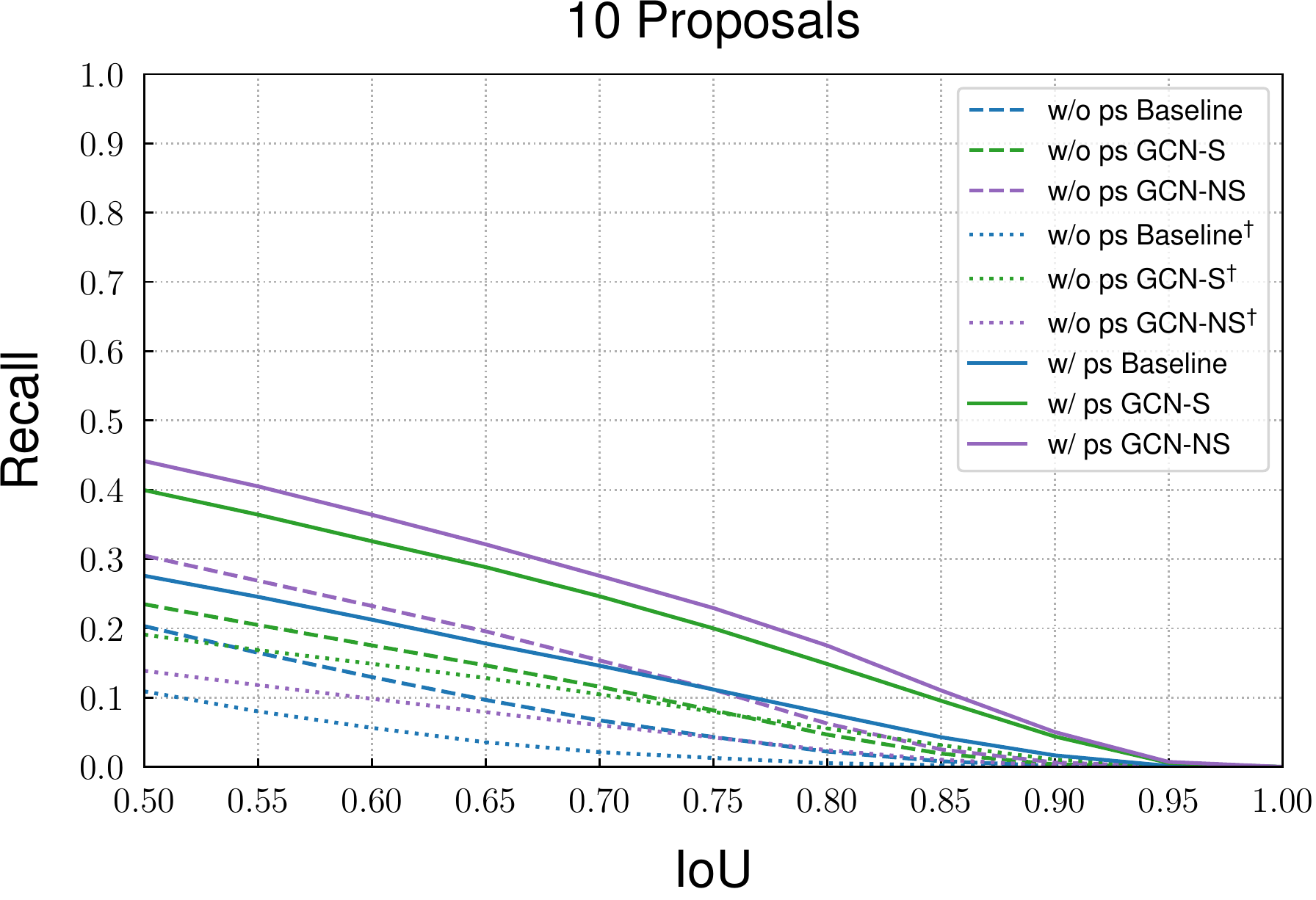}
	\endminipage\hfill
	\minipage{0.25\textwidth}
	\includegraphics[width=\linewidth]{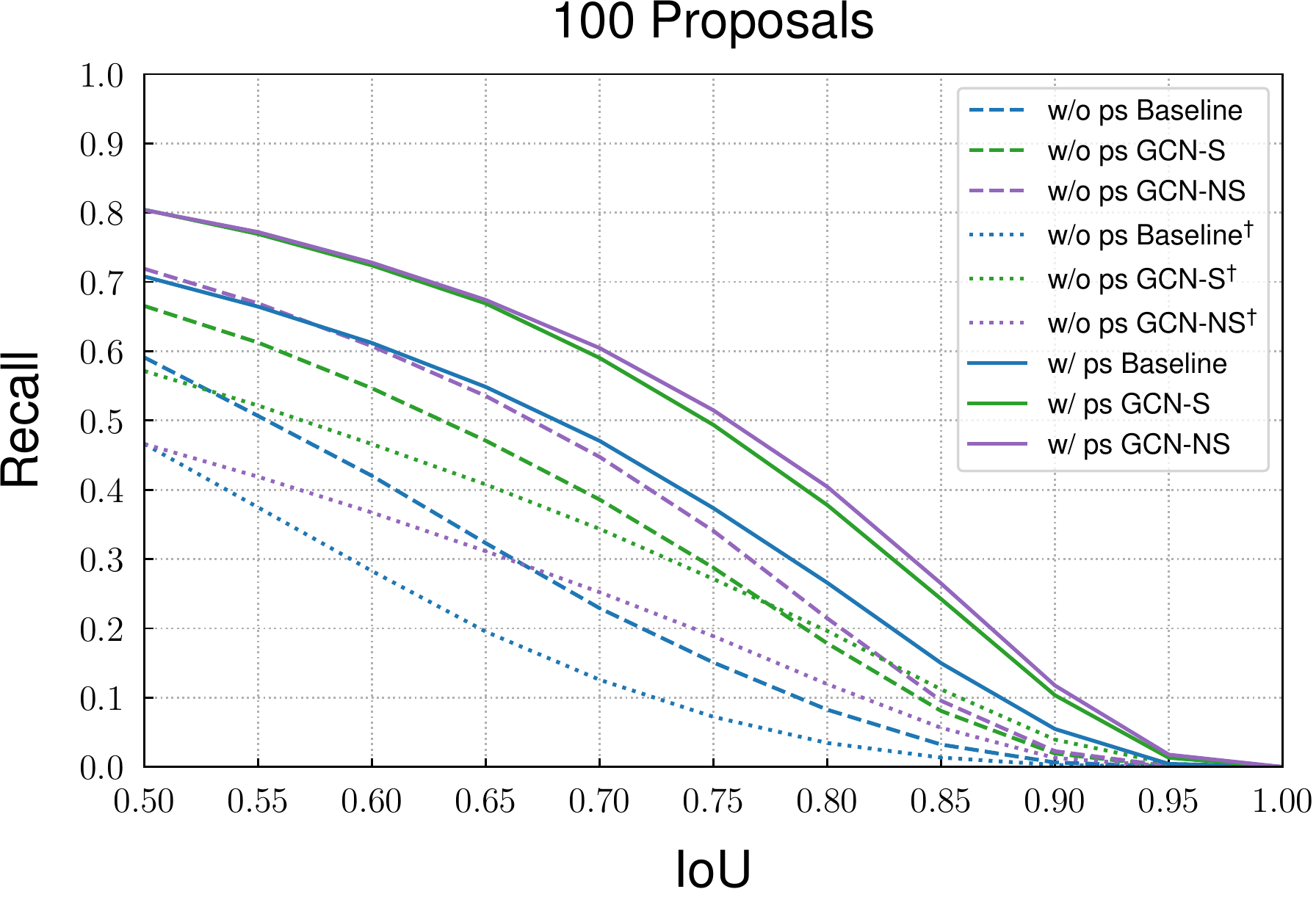}
	\endminipage\hfill
	\minipage{0.25\textwidth}
	\includegraphics[width=\linewidth]{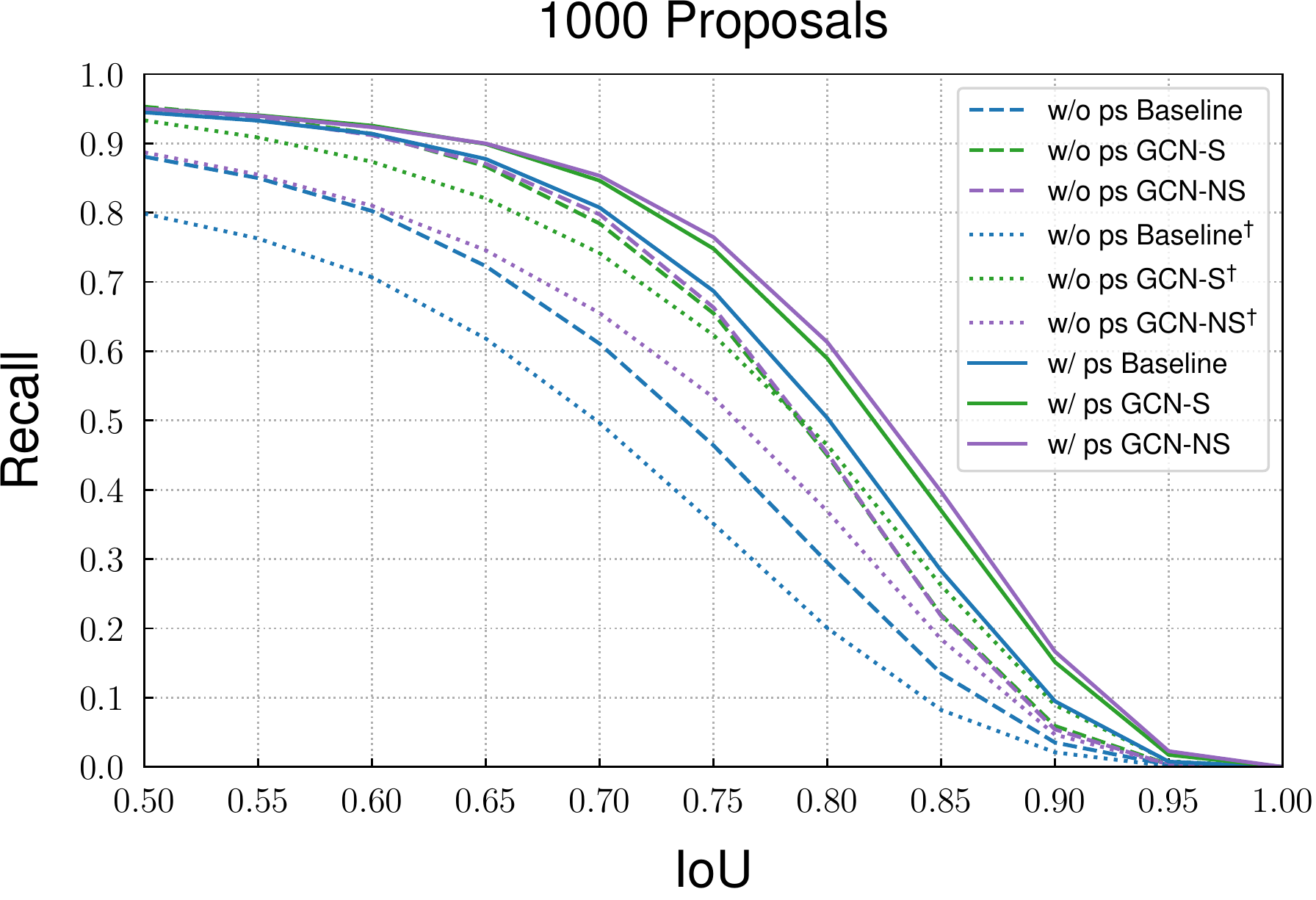}
	\endminipage\hfill
	\minipage{0.25\textwidth}
	\includegraphics[width=\linewidth]{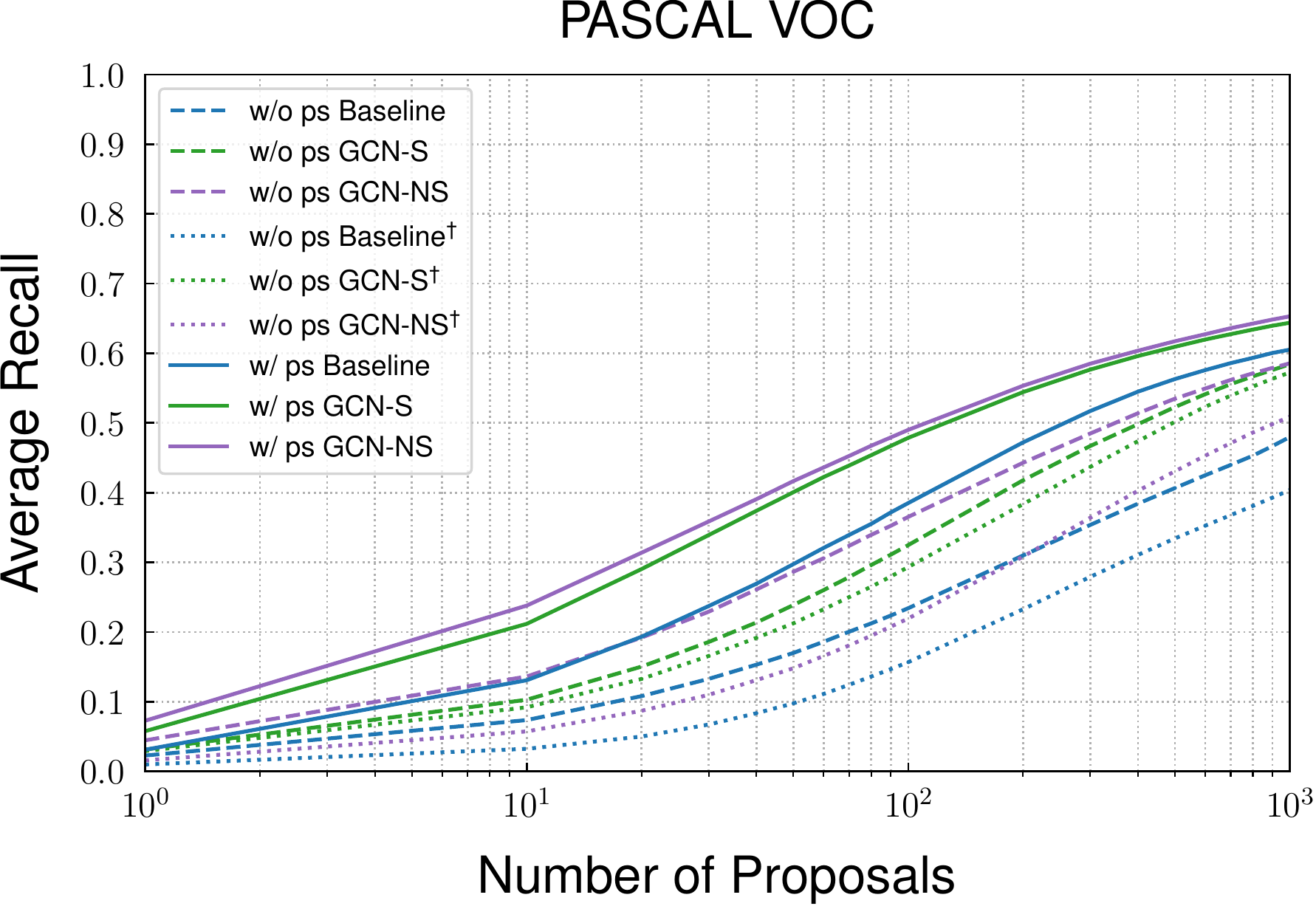}
	\endminipage
	\caption{Recall against IoU with different proposal numbers of 10, 100, 1,000 and average recall against the number of proposals on PASCAL VOC \texttt{07test} using models trained on PASCAL VOC \texttt{07+12}. Note that models with \textsuperscript{\textdagger} use 5 different anchor aspect ratios to generate more anchors.}
	\label{fig:recall-nprop}
\end{figure}

\begin{figure}[t]
	\centering
	\minipage{0.25\textwidth}
	\includegraphics[width=\linewidth]{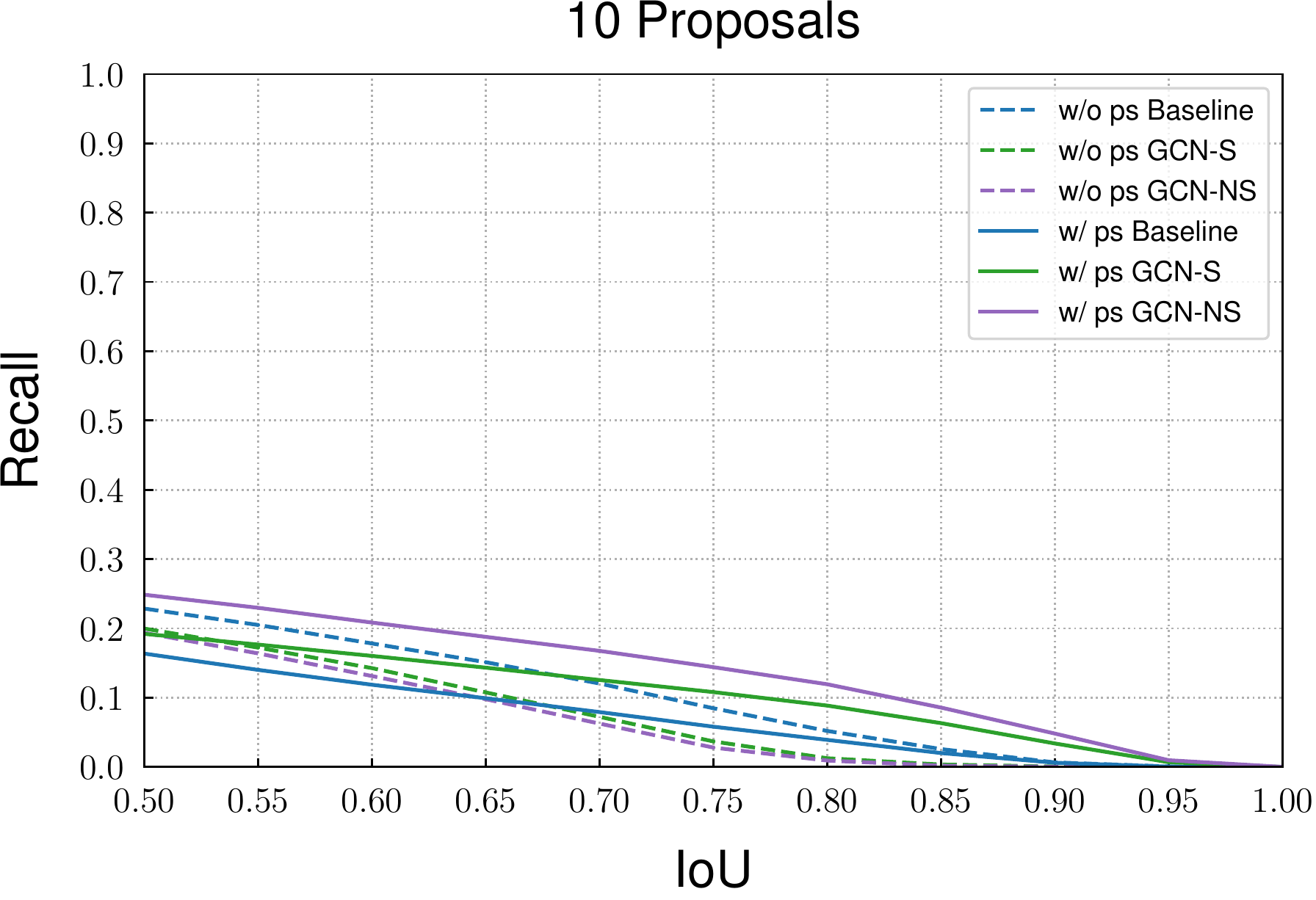}
	\endminipage\hfill
	\minipage{0.25\textwidth}
	\includegraphics[width=\linewidth]{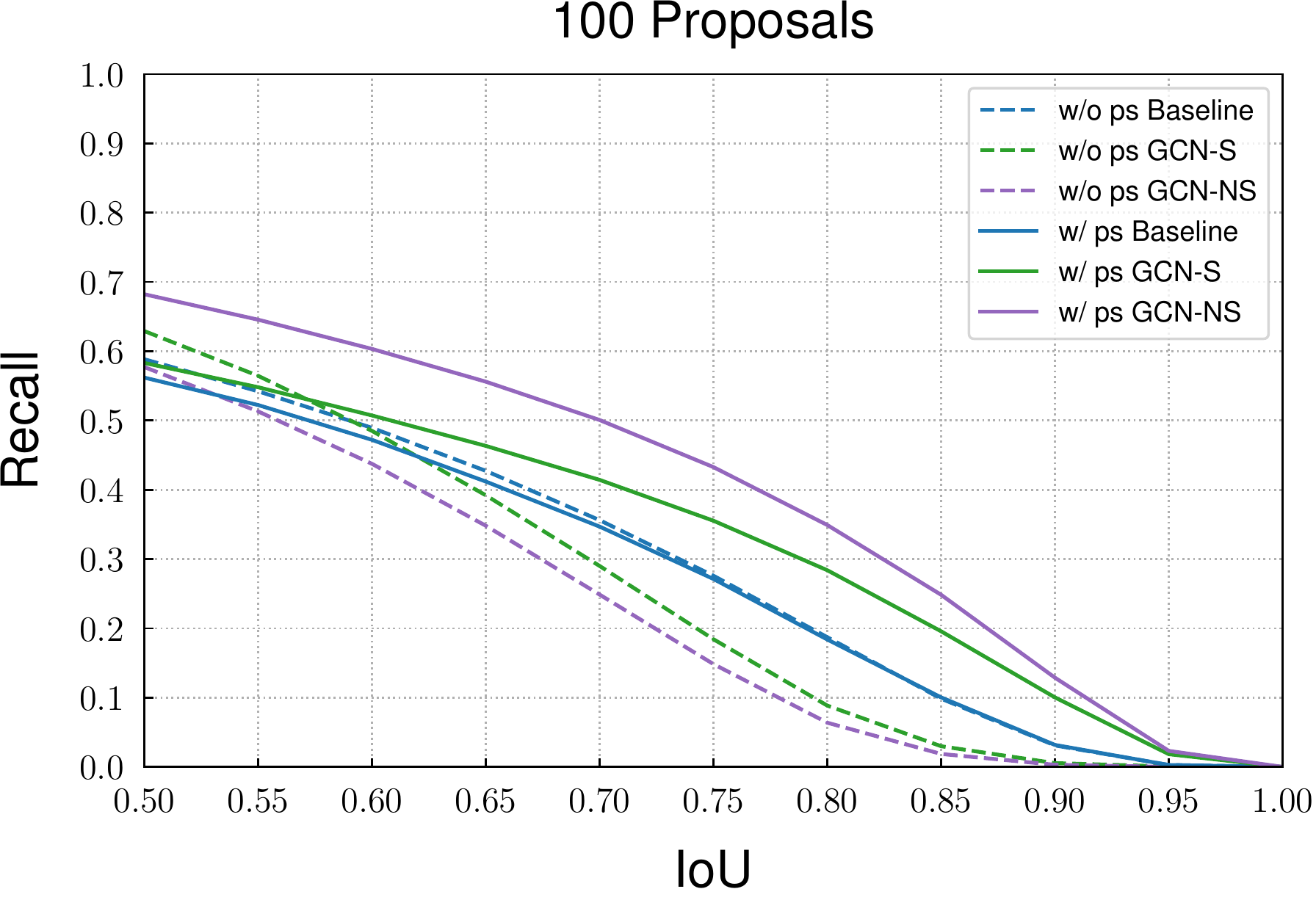}
	\endminipage\hfill
	\minipage{0.25\textwidth}%
	\includegraphics[width=\linewidth]{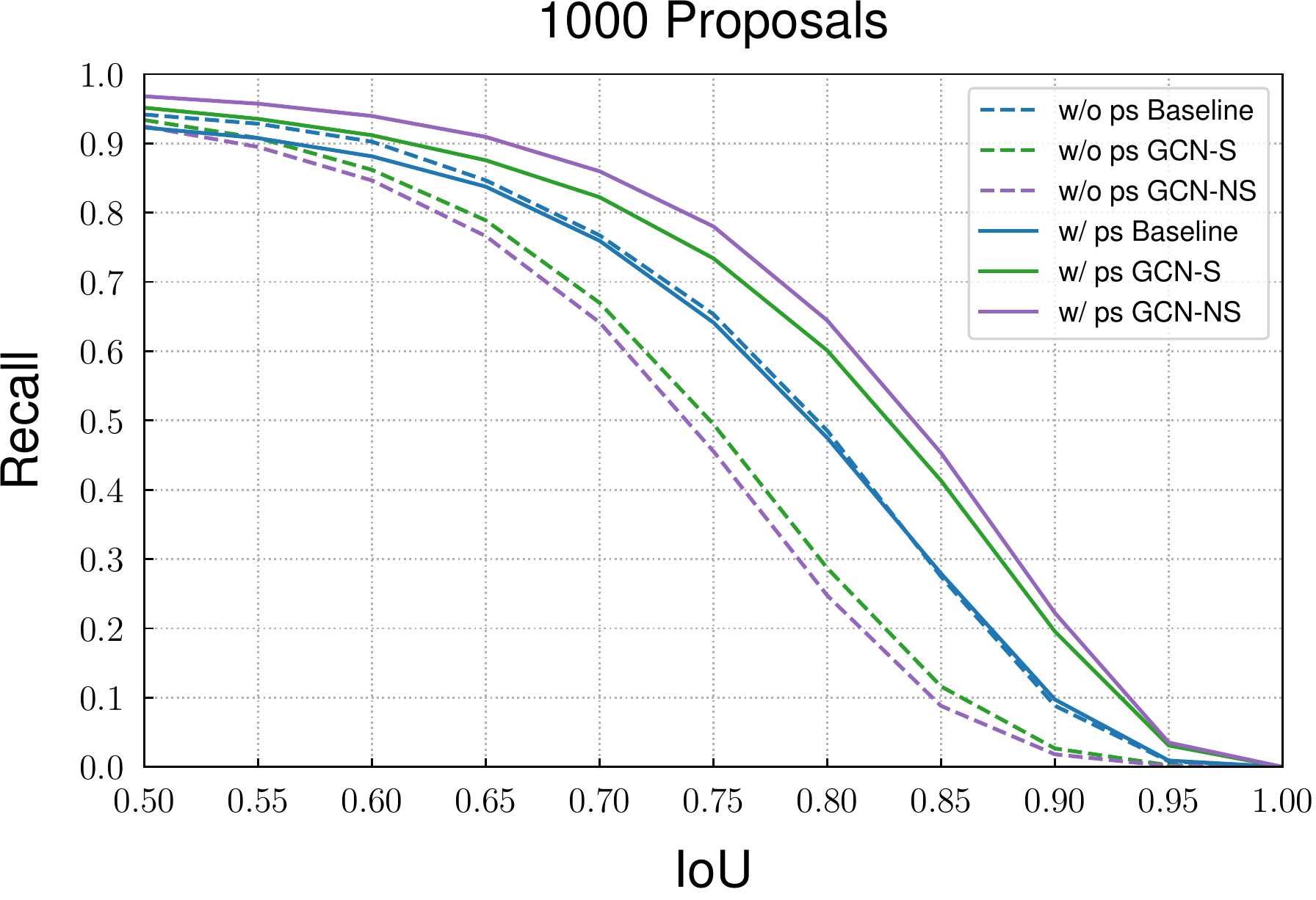}
	\endminipage\hfill
	\minipage{0.25\textwidth}%
	\includegraphics[width=\linewidth]{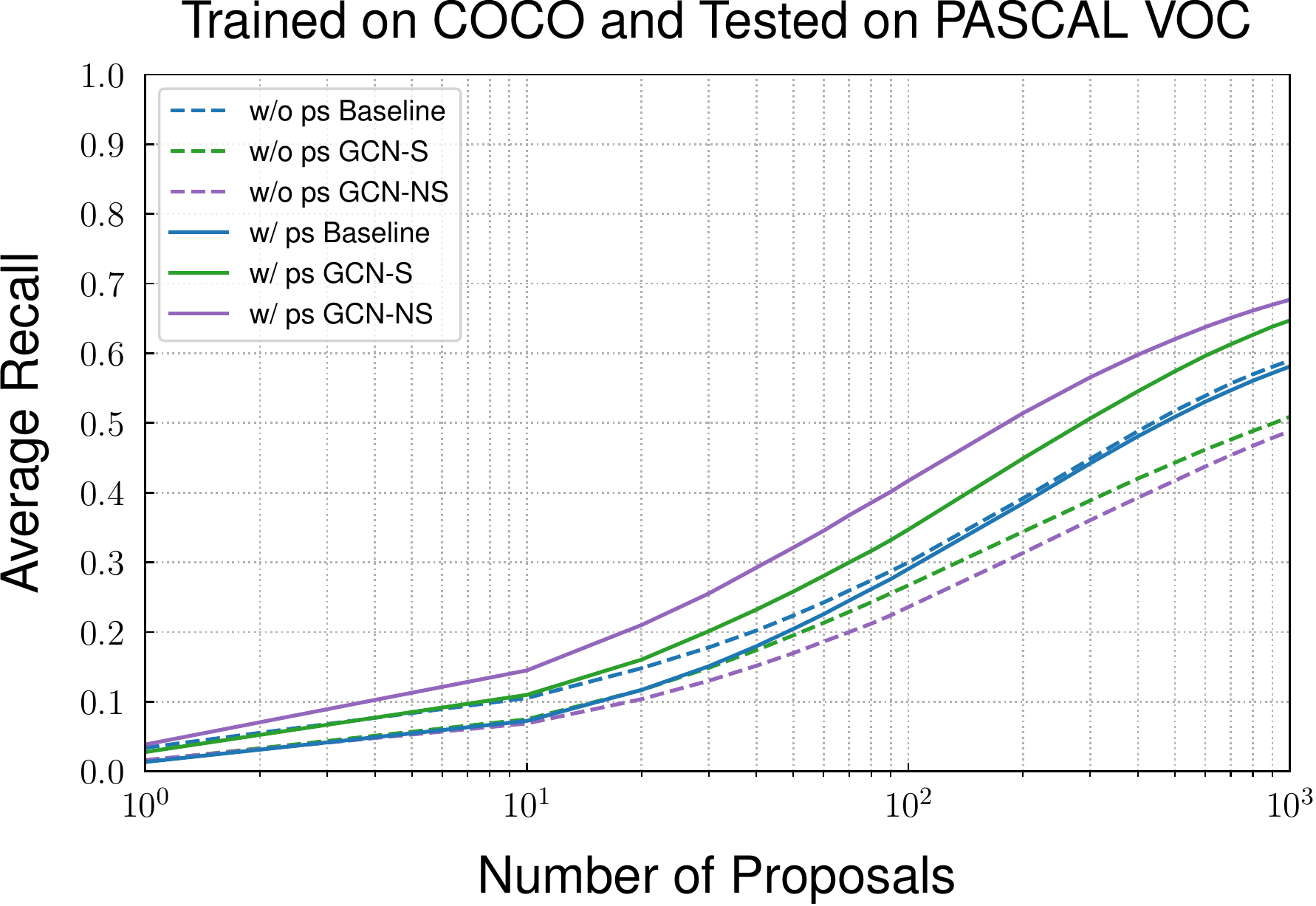}
	\endminipage
	\caption{Recall against IoU with different proposal numbers of 10, 100 and 1,000 and average recall against the number of proposals on PASCAL VOC \texttt{07test} using models trained on COCO \texttt{trainval35k}.}
	\label{fig:recall-iou-nprop-supp}
\end{figure}

\begin{figure}[t]
	\centering
	\includegraphics[width=0.95\textwidth]{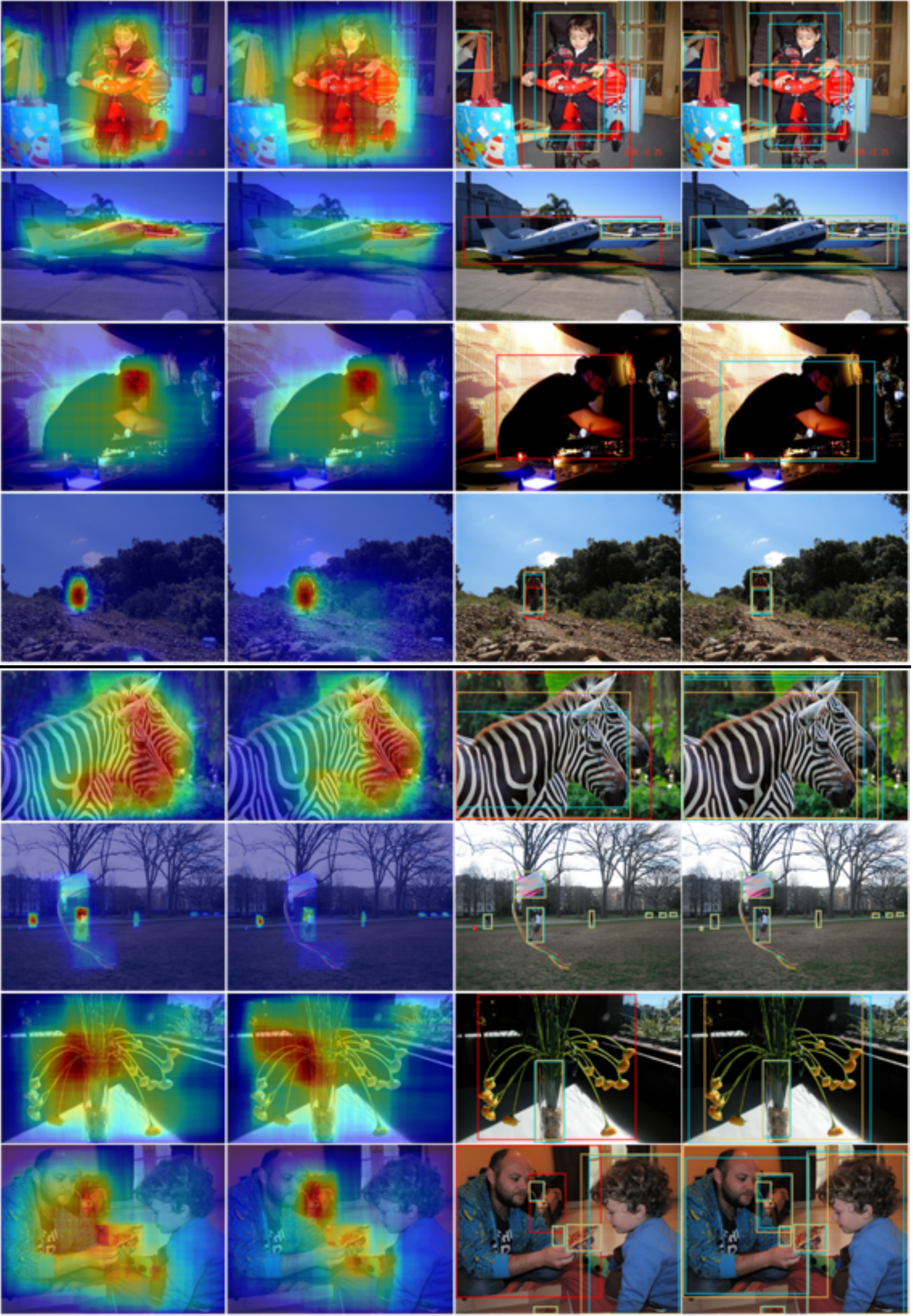}
	\caption{The impact of position-sensitivity: visualisation on the distribution heatmap and hit-and-miss at the top 1,000 proposals by GCN-NS models for a number of selected PASCAL VOC \texttt{07test} (\textbf{Row 1-4}) and COCO \texttt{minival} (\textbf{Row 5-8}) images. For each pair of images, model without position-sensitivity is on the left and the one with position-sensitivity is on the right. \textbf{Col 1-2:} The heatmaps are plotted by stacking the proposal boxes. \textbf{Col 3-4:} The bounding boxes in orange are ground truth boxes with their corresponding hit proposals in cyan, in which the IoU threshold is set to 0.7, and the bounding boxes in red are missed cases.}
	\label{fig:hit-miss-exp-supp}
\end{figure}

\begin{figure}[t]
	\centering
	\includegraphics[width=0.95\textwidth]{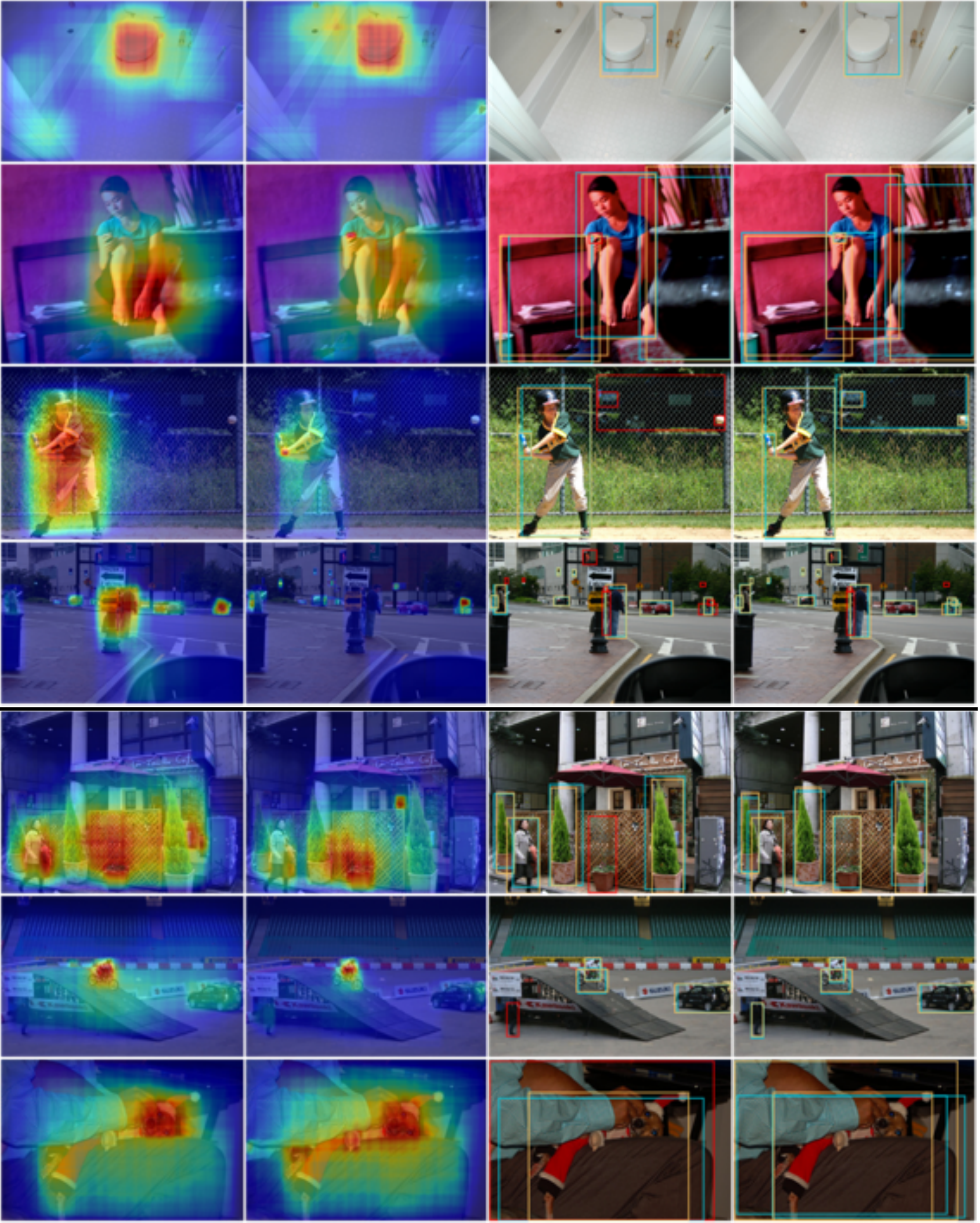}
	\caption{The results of model generalisation and capacity experiments visualised in the distribution heatmap and hit-and-miss of the top 1,000 proposals by GCN-NS models for a number of selected COCO \texttt{minival} (\textbf{Row 1-4}) and PASCAL VOC \texttt{07test} (\textbf{Row 5-7}) images. 
		\textbf{Row 1-4:} For each pair of COCO \texttt{minival} images, the result of model trained on PASCAL VOC is on the left and the one of model trained on COCO is on the right.
		\textbf{Row 5-7:} For each pair of PASCAL VOC \texttt{07test} images, the result of model trained on PASCAL VOC is on the left and the one of model trained on COCO is on the right.}
	\label{fig:generalisation-hit-miss-exp-supp}
\end{figure}

In this supplementary material, we present more experiment results to complement the manuscript.
\begin{itemize}
	\item The number of anchors generated by non-window (i.e., the original RPN method \cite{Ren2015}) and window mapping can end up being very different, depending on the used anchor profile. The proposed window mapping method using three aspect ratios~\cite{Lin2017} increases about twice in the number of anchors compared with the non-window mapping method. One may argue that the improvement of the proposed method was due to the large number of anchors. To investigate the genuine impact of increasing anchor number, we added two more anchor aspect ratios \{$1:3$, $3:1$\} based on the original profile (i.e., \{$1:1$, $1:2$, $2:1$\}), which in total forms 5 different anchors, for the models without using position-sensitivity. Table~\ref{tab:anchor-number} shows the anchor number for window mapping and non-window using three profiles and five profiles. 
	The total number of anchors and the results of average recall (AR) of the models (5-anchor models marked with \textsuperscript{\textdagger}) are reported in Table~\ref{tab:pascal-results-supp} and Fig.~\ref{fig:recall-nprop}. As one can see that simply increasing the number of different anchor profiles, and thus to increase the total number of anchors, does not necessarily improve the overall AR. This means that the position-sensitive score maps can be an harmonious adoption for the anchors generated by the window mapping method.
	
	\item We conducted a model capacity experiment by training a model on COCO \texttt{trainval35k} and test it on PASCAL VOC \texttt{07test}. The results are shown in Table~\ref{tab:coco-pascal-results} and Fig.~\ref{fig:recall-iou-nprop-supp}. Compared to the models trained on PASCAL VOC \texttt{07+12}, the GCN-NS models trained on COCO \texttt{trainval35k} can achieve even higher $\textrm{AR}^{1k}$ scores from $0.653$ (see Table~\ref{tab:pascal-results} in main paper) to $0.678$ (4\%). The overall improvement can be broken down into the boost in $\textrm{AR}^{1k}_{s}$, $\textrm{AR}^{1k}_{m}$ and $\textrm{AR}^{1k}_{l}$ due to more data with different object sizes in COCO. This indicates that the AR results of the proposed method can still be improved by using more bounding box training data. Fig.~\ref{fig:generalisation-hit-miss-exp-supp} (\textbf{Row 5-7}) shows the corresponding examples. This observation corroborates the assumption that given enough objects from various categories, the classifier can be generalised to capture the semantic meaning of objectness, which aligns with precious works~\cite{Chavali2016,Hosang2016,Kuo2015,Pinheiro2015}.  
	
	\item Fig.~\ref{fig:hit-miss-exp-supp} displays more examples of the distribution heatmap and hit-and-miss at top 1,000 proposals using GCN-NS models with and without taking position-sensitivity into account. The ground truth IoU threshold is set to 0.7, the same as in the manuscript. 
	
	\item Fig.~\ref{fig:generalisation-hit-miss-exp-supp} shows more examples to display model capacity and model generalisation of our proposed method.
\end{itemize}
\begin{table}
	\begin{minipage}[t]{\linewidth} 
		\centering
		\caption{The numbers of anchors in each layer with respect to different anchor profiles with fixed input size $640\times 640$. Note that the default non-window model uses 3 anchor profiles, and models with \textsuperscript{\textdagger} use 5 different anchor aspect ratios to generate more anchors}
		\label{tab:anchor-number}
		\begin{tabular}{rrrr}
			\toprule
			& non-window & non-window\textsuperscript{\textdagger} & window \\
			\midrule
			$D_2$ & 76,800 & 128,000 &  194,058\\
			$D_3$ & 19,200 & 32,000  &27,803\\
			$D_4$ & 4,800  & 8,000	 &5,963 \\
			$D_5$ & 1,200  & 2,000	 &1,043\\
			$D_6$ & 300    & 500	 &83\\
			\textbf{Total} & 102,300 & 170,500 &228,950 \\
			\bottomrule
		\end{tabular}
	\end{minipage}%
\end{table}
\begin{table}
	\begin{minipage}[t]{\linewidth} 
		\caption{Object proposal results of all models trained on PASCAL VOC \texttt{07+12} and evaluated on \texttt{07test} using different anchor profiles}
		\label{tab:pascal-results-supp}
		\centering
		\resizebox{\textwidth}{!} {
			\begin{tabular}{lllllllllllllllll}
				\toprule
				\multicolumn{2}{c}{} & 
				\multicolumn{7}{c}{3-anchor w/o position-sensitive} & \multicolumn{1}{c}{} & 
				\multicolumn{7}{c}{5-anchor w/o position-sensitive\textsuperscript{\textdagger}} \\
				\cmidrule(lr){2-9}
				\cmidrule(lr){10-17}
				\multicolumn{2}{c}{} &
				\multicolumn{1}{c}{$\textrm{AR}^{10}$} &
				\multicolumn{1}{c}{$\textrm{AR}^{100}$} &
				\multicolumn{1}{c}{$\textrm{AR}^{1k}$} &
				\multicolumn{1}{c}{$\textrm{AUC}$} &
				\multicolumn{1}{c}{$\textrm{AR}_{s}^{1k}$} &
				\multicolumn{1}{c}{$\textrm{AR}_{m}^{1k}$} &
				\multicolumn{1}{c}{$\textrm{AR}_{l}^{1k}$} &
				\multicolumn{1}{c}{} &
				\multicolumn{1}{c}{$\textrm{AR}^{10}$} &
				\multicolumn{1}{c}{$\textrm{AR}^{100}$} &
				\multicolumn{1}{c}{$\textrm{AR}^{1k}$} &
				\multicolumn{1}{c}{$\textrm{AUC}$} &
				\multicolumn{1}{c}{$\textrm{AR}_{s}^{1k}$} &
				\multicolumn{1}{c}{$\textrm{AR}_{m}^{1k}$} &
				\multicolumn{1}{c}{$\textrm{AR}_{l}^{1k}$}
				\\
				\midrule
				Baseline 	& & .074 & .234 & .480 & .272 & .254 & .414 & .566 & & .032 & .157 & .404 & .204 & .200 & .252 & .539 \\ \midrule
				GCN-S		& & .103 & .325 & .584 & .356 & \underline{\textbf{.471}} & .558 & .623 & & \underline{.092} & \underline{.293} & \underline{.572} & \underline{.335} & \underline{.451} & \underline{.553} & \underline{.609} \\
				GCN-NS 		& & \underline{\textbf{.136}} & \underline{\textbf{.365}} & \underline{\textbf{.586}} & \underline{\textbf{.383}} & .445 & \underline{\textbf{.569}} & \underline{\textbf{.625}} & & .057 & .219 & .509 & .272 & .442 & .516 & .518 \\
				\bottomrule
			\end{tabular}%
		}
	\end{minipage}
\end{table}
\begin{table}[t]
	\begin{minipage}[t]{\linewidth} 
		\caption{Object proposal results of all models trained on COCO \texttt{trainval35k} and evaluated on PASCAL VOC \texttt{07test}}
		\label{tab:coco-pascal-results}
		\centering
		\resizebox{\textwidth}{!} {
			\begin{tabular}{lllllllllllllllll}
				\toprule
				\multicolumn{2}{c}{} & 
				\multicolumn{7}{c}{w/o position-sensitive} & \multicolumn{1}{c}{} & 
				\multicolumn{7}{c}{w/ position-sensitive} \\
				\cmidrule(lr){2-9}
				\cmidrule(lr){10-17}
				\multicolumn{2}{c}{} &
				\multicolumn{1}{c}{$\textrm{AR}^{10}$} &
				\multicolumn{1}{c}{$\textrm{AR}^{100}$} &
				\multicolumn{1}{c}{$\textrm{AR}^{1k}$} &
				\multicolumn{1}{c}{$\textrm{AUC}$} &
				\multicolumn{1}{c}{$\textrm{AR}_{s}^{1k}$} &
				\multicolumn{1}{c}{$\textrm{AR}_{m}^{1k}$} &
				\multicolumn{1}{c}{$\textrm{AR}_{l}^{1k}$} &
				\multicolumn{1}{c}{} &
				\multicolumn{1}{c}{$\textrm{AR}^{10}$} &
				\multicolumn{1}{c}{$\textrm{AR}^{100}$} &
				\multicolumn{1}{c}{$\textrm{AR}^{1k}$} &
				\multicolumn{1}{c}{$\textrm{AUC}$} &
				\multicolumn{1}{c}{$\textrm{AR}_{s}^{1k}$} &
				\multicolumn{1}{c}{$\textrm{AR}_{m}^{1k}$} &
				\multicolumn{1}{c}{$\textrm{AR}_{l}^{1k}$}
				\\
				\midrule
				Baseline 	& & .108 & .301 & .605 & .352 & .442 & \underline{.607} & .636 & & .054 & .248 & .564 & .301 & .499 & .501 & .616 \\ \midrule
				GCN-S		& & \underline{.134} & \underline{.407} & \underline{.666} & \underline{.427} & .513 & .604 & \underline{.735} & & .119 & .348 & .648 & .390 & .538 & .595 & .703 \\
				GCN-NS 		& & .122 & .376 & .651 & .404 & \underline{.518} & .602 & .707 & & \underline{\textbf{.158}} & \underline{\textbf{.410}} & \underline{\textbf{.678}} & \underline{\textbf{.437}} & \underline{\textbf{.550}} & \underline{\textbf{.625}} & \underline{\textbf{.736}} \\
				\bottomrule
			\end{tabular}%
		}
	\end{minipage}
\end{table}
\clearpage
\bibliographystyle{splncs04}

\begin{thebibliography}{10}
\providecommand{\url}[1]{\texttt{#1}}
\providecommand{\urlprefix}{URL }
\providecommand{\doi}[1]{https://doi.org/#1}

\bibitem{Arbelaez2014}
Arbel{\'a}ez, P., Pont-Tuset, J., Barron, J.T., Marques, F., Malik, J.:
  Multiscale combinatorial grouping. In: CVPR (2014)

\bibitem{Bell2016}
Bell, S., Lawrence~Zitnick, C., Bala, K., Girshick, R.: Inside-outside net:
  Detecting objects in context with skip pooling and recurrent neural networks.
  In: CVPR (2016)

\bibitem{Carreira2012}
Carreira, J., Sminchisescu, C.: Cpmc: Automatic object segmentation using
  constrained parametric min-cuts. In: TPAMI (2012)

\bibitem{Chavali2016}
Chavali, N., Agrawal, H., Mahendru, A., Batra, D.: Object-proposal evaluation
  protocol is 'gameable'. In: CVPR (2016)

\bibitem{Dai2016}
Dai, J., Li, Y., He, K., Sun, J.: {R-FCN}: Object detection via region-based
  fully convolutional networks. In: NIPS (2016)

\bibitem{Everingham2010}
Everingham, M., Van~Gool, L., Williams, C.K., Winn, J., Zisserman, A.: The
  pascal visual object classes ({VOC}) challenge. In: IJCV (2010)

\bibitem{Felzenszwalb2010}
Felzenszwalb, P.F., Girshick, R.B., McAllester, D., Ramanan, D.: Object
  detection with discriminatively trained part-based models. In: TPAMI (2010)

\bibitem{Gidaris2015}
Gidaris, S., Komodakis, N.: Object detection via a multi-region and semantic
  segmentation-aware cnn model. In: ICCV (2015)

\bibitem{Girshick2015}
Girshick, R.: Fast r-cnn. In: ICCV (2015)

\bibitem{Girshick2014}
Girshick, R., Donahue, J., Darrell, T., Malik, J.: Rich feature hierarchies for
  accurate object detection and semantic segmentation. In: CVPR (2014)

\bibitem{He2014}
He, K., Zhang, X., Ren, S., Sun, J.: Spatial pyramid pooling in deep
  convolutional networks for visual recognition. In: ECCV (2014)

\bibitem{He2016}
He, K., Zhang, X., Ren, S., Sun, J.: Deep residual learning for image
  recognition. In: CVPR (2016)

\bibitem{Hosang2016}
Hosang, J., Benenson, R., Doll{\'a}r, P., Schiele, B.: What makes for effective
  detection proposals? In: TPAMI (2016)

\bibitem{Kong2016}
Kong, T., Yao, A., Chen, Y., Sun, F.: Hypernet: Towards accurate region
  proposal generation and joint object detection. In: CVPR (2016)

\bibitem{Kuo2015}
Kuo, W., Hariharan, B., Malik, J.: {DeepBox}: Learning objectness with
  convolutional networks. In: ICCV (2015)

\bibitem{Lin2017}
Lin, T.Y., Doll{\'a}r, P., Girshick, R., He, K., Hariharan, B., Belongie, S.:
  Feature pyramid networks for object detection. In: CVPR (2017)

\bibitem{Lin2014}
Lin, T.Y., Maire, M., Belongie, S., Hays, J., Perona, P., Ramanan, D.,
  Doll{\'a}r, P., Zitnick, C.L.: {Microsoft COCO}: Common objects in context.
  In: ECCV (2014)

\bibitem{Long2015}
Long, J., Shelhamer, E., Darrell, T.: Fully convolutional networks for semantic
  segmentation. In: CVPR (2015)

\bibitem{Peng2017}
Peng, C., Zhang, X., Yu, G., Luo, G., Sun, J.: Large kernel matters--improve
  semantic segmentation by global convolutional network. In: CVPR (2017)

\bibitem{Pinheiro2015}
Pinheiro, P.O., Collobert, R., Doll{\'a}r, P.: Learning to segment object
  candidates. In: NIPS (2015)

\bibitem{Pinheiro2016}
Pinheiro, P.O., Lin, T.Y., Collobert, R., Doll{\'a}r, P.: Learning to refine
  object segments. In: ECCV (2016)

\bibitem{Pont2015}
Pont-Tuset, J., Van~Gool, L.: Boosting object proposals: From pascal to coco.
  In: ICCV (2015)

\bibitem{Ren2015}
Ren, S., He, K., Girshick, R., Sun, J.: Faster r-cnn: Towards real-time object
  detection with region proposal networks. In: Advances in neural information
  processing systems (2015)

\bibitem{Ronneberger2015}
Ronneberger, O., Fischer, P., Brox, T.: U-net: Convolutional networks for
  biomedical image segmentation. In: MICCAI (2015)

\bibitem{Shrivastava2016}
Shrivastava, A., Gupta, A., Girshick, R.: Training region-based object
  detectors with online hard example mining. In: CVPR (2016)

\bibitem{Simonyan2014}
Simonyan, K., Zisserman, A.: Very deep convolutional networks for large-scale
  image recognition. In: ICLR (2014)

\bibitem{Uijlings2013}
Uijlings, J.R., Van De~Sande, K.E., Gevers, T., Smeulders, A.W.: Selective
  search for object recognition. In: IJCV (2013)

\bibitem{Viola2004}
Viola, P., Jones, M.J.: Robust real-time face detection. In: IJCV (2004)

\bibitem{Zeiler2014}
Zeiler, M.D., Fergus, R.: Visualizing and understanding convolutional networks.
  In: ECCV (2014)

\bibitem{Zitnick2014}
Zitnick, C.L., Doll{\'a}r, P.: Edge boxes: Locating object proposals from
  edges. In: ECCV (2014)

\end{thebibliography}

\begin{thebibliography}{1}
\providecommand{\url}[1]{\texttt{#1}}
\providecommand{\urlprefix}{URL }
\providecommand{\doi}[1]{https://doi.org/#1}

\bibitem{Chavali2016}
Chavali, N., Agrawal, H., Mahendru, A., Batra, D.: Object-proposal evaluation
  protocol is 'gameable'. In: CVPR (2016)

\bibitem{Hosang2016}
Hosang, J., Benenson, R., Doll{\'a}r, P., Schiele, B.: What makes for effective
  detection proposals? In: TPAMI (2016)

\bibitem{Kuo2015}
Kuo, W., Hariharan, B., Malik, J.: {DeepBox}: Learning objectness with
  convolutional networks. In: ICCV (2015)

\bibitem{Lin2017}
Lin, T.Y., Doll{\'a}r, P., Girshick, R., He, K., Hariharan, B., Belongie, S.:
  Feature pyramid networks for object detection. In: CVPR (2017)

\bibitem{Pinheiro2015}
Pinheiro, P.O., Collobert, R., Doll{\'a}r, P.: Learning to segment object
  candidates. In: NIPS (2015)

\bibitem{Ren2015}
Ren, S., He, K., Girshick, R., Sun, J.: Faster r-cnn: Towards real-time object
  detection with region proposal networks. In: Advances in neural information
  processing systems (2015)

\end{thebibliography}

\end{document}